\documentclass[twoside]{article}

\usepackage[accepted]{aistats2026}

\usepackage{gav_style}
\usepackage{natbib}
\usepackage{graphicx}

\usepackage[table]{xcolor}

\usepackage[utf8]{inputenc} %
\usepackage[T1]{fontenc}    %
\usepackage{hyperref}       %
\usepackage{url}            %
\usepackage{booktabs}       %
\usepackage{amsfonts}       %
\usepackage{nicefrac}       %
\usepackage{microtype}      %
\usepackage{xcolor}         %

\usepackage{thmtools} 
\usepackage{layouts}
\usepackage{placeins}
\usepackage{wrapfig}

\SetKwComment{Comment}{/* }{ */}

\usepackage{xr-hyper}      %

\begin{document}

\twocolumn[

\aistatstitle{EventFlow: Forecasting Temporal Point Processes with Flow Matching}

\aistatsauthor{ Gavin Kerrigan$^\dagger$ \And Kai Nelson$^\dagger$ \And  Padhraic Smyth }

\aistatsaddress{ University of Oxford \\ \texttt{kerrigan@stats.ox.ac.uk} \And  University of California, Berkeley \\ \texttt{kai\_nelson@berkeley.edu} \And University of California, Irvine \\ \texttt{smyth@ics.uci.edu} }

]

\begin{abstract}
  Continuous-time event sequences, in which events occur at irregular intervals, are ubiquitous across a wide range of industrial and scientific domains. The contemporary modeling paradigm is to treat such data as realizations of a temporal point process, and in machine learning it is common to model temporal point processes in an autoregressive fashion using a neural network. While autoregressive models are successful in predicting the time of a single subsequent event, their performance can degrade when forecasting longer horizons due to cascading errors and myopic predictions. We propose \texttt{EventFlow}, a non-autoregressive generative model for temporal point processes. The model builds on the flow matching framework in order to directly learn joint distributions over event times, side-stepping the autoregressive process. \texttt{EventFlow} is simple to implement and achieves a 20\%-53\% lower forecast error than the nearest baseline on standard TPP benchmarks while simultaneously using fewer model calls at sampling time. %
\end{abstract}

\section{Introduction}

\looseness=-1
Many stochastic processes, ranging from the occurrence of earthquakes \citep{ogata1998space} to consumer behavior \citep{xu2014path}, are best understood as a sequence of discrete events that occur at random times. Any observed event sequence, consisting of one or more event times, may be viewed as a draw from a temporal point process (TPP) \citep{daley2003introduction} which characterizes the distribution over such sequences. Given a collection of observed event sequences, faithfully modeling the underlying TPP is critical in both understanding and forecasting the phenomenon of interest. %

\looseness=-1
While multiple different parametric TPP models have been proposed \citep{hawkes1971spectra, isham1979self}, their limited flexibility limits their application when modeling complex real-world sequences. This has motivated the use of neural networks \citep{du2016recurrent, mei2017neural} in modeling TPPs. To date, most neural TPP models are autoregressive in nature \citep{shchurintensity, zhang2020self, shchur2021tppreview}, where a model is trained to predict only a single subsequent event time given an observed history of events. However, in many tasks, we are interested not only in the next event, but in the \textit{entire sequence of events }which is to follow. While autoregressive neural TPP models can be applied in this setting, their performance in many-step forecasting tasks can be unsatisfactory due to compounding errors arising from the autoregressive sampling procedure \citep{xue2022hypro, ludke2023add}. %

Moreover, existing models are typically trained via a maximum likelihood procedure (see Section \ref{sec:background}) which involves computing the CDF implied by the learned model. When using a neural model, computing this CDF often requires techniques such as Monte Carlo estimation to compute the loss \citep{mei2017neural}. In addition, sampling from intensity-based models \citep{du2016recurrent, mei2017neural, mei2022transformer} is nontrivial, requiring an expensive and difficult to implement approach based on the thinning algorithm \citep{lewis1979simulation, ogata1981lewis, xue2023easytpp}.

\begin{figure*}[!t]
    \centering
    \includegraphics[width=4.8in, height=1.75in]{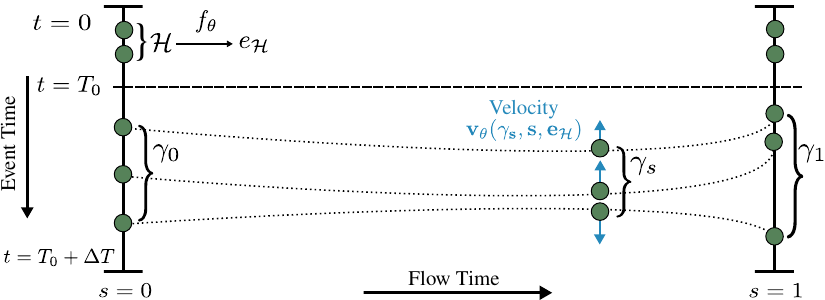}
    \caption{All illustration of forecasting with our \texttt{EventFlow} method. The horizontal axis indicates the flow time $s$, and the vertical axis indicates the support of the TPP $\TT = [0, T]$. We first encode the observed history $\HH$ into an embedding $e_{\HH} = f_\theta(\HH)$. At $s=0$, we independently draw $n$ events in the forecasting window $[T_0, T_0 + \Delta T]$ from a fixed reference distribution, constituting a sample $\gamma_0$ from a mixed-binomial TPP. Each event can be thought of as a particle, which is assigned a velocity by a neural network $v_\theta(\gamma_s, s, e_{\HH})$. Each particle flows along its velocity field until reaching its terminal point at $s=1$, whereby we obtain a forecasted sequence $\gamma_1$.}
    \label{fig:leading_figure}
\end{figure*}

\looseness=-1
Motivated by these limitations, we propose \texttt{EventFlow}, a generative model which directly learns the full joint distribution over future event times, allowing us to avoid autoregressive sampling altogether. Our proposed model extends the flow matching framework \citep{lipman2023flow, albergo2022building, liu2022flow} to the setting of TPPs. 
To generate a sample, we first draw a set of random event times from a simple reference distribution and then transport them through a learned vector field to obtain a sequence of predicted times. The number of events is held fixed during this transformation, decoupling the modeling of event counts from that of event times. See Figure \ref{fig:leading_figure} for an illustration.

This yields a flexible formulation that can be used for both unconditional generation (sampling from the underlying TPP) and conditional generation (forecasting future events given a history). More broadly, \texttt{EventFlow} provides an alternative perspective on TPP modeling: rather than specifying an intensity function and relying on sequential simulation, we decompose the generative process into a distribution over event counts and a flow-based model over event times. This results in a non-autoregressive model that is straightforward to train and efficient to sample from.

Empirically, we demonstrate that \texttt{EventFlow} obtains state-of-the-art performance on multi-step forecasting benchmarks and is competitive with leading approaches for unconditional generation. In particular, our model reduces forecast error by 20\%–53\% relative to the strongest baselines. Further, compared with existing methods, which can require hundreds of model evaluations at sampling time, we demonstrate that \texttt{EventFlow} can achieve competitive performance with only a \textit{single} forward pass of the underlying neural network.

\section{Related Work}
\label{sec:related_work}

\looseness=-1
\paragraph{Temporal Point Processes} The statistical modeling of temporal point processes (TPPs) has a long history \citep{daley2003introduction, hawkes1971spectra, isham1979self}. The contemporary modeling paradigm, based on neural networks \citep{du2016recurrent, shchur2021tppreview}, typically operates by learning a \textit{history encoder} and an \textit{event decoder}. The history encoder seeks to learn a fixed-dimensional vector representation of the sequence of events which have been observed or already predicted, and the decoder seeks to model a distribution over the subsequent event time(s). %

\looseness=-1
Popular choices for the history encoder include RNN-based models \citep{du2016recurrent, shchurintensity, mei2019imputing} or attention-based models \citep{zhang2020self, zuo2020transformer, mei2022transformer}. While attention-based encoders can provide longer-range contexts, this benefit typically comes at the cost of additional memory overhead. %

\looseness=-1
For the event decoder, a common approach is to parametrize a conditional intensity function using a neural network that takes as input a learned representation of the event history \citep{du2016recurrent, mei2017neural, zuo2020transformer}. An alternative and more recent approach is to use generative models as decoders. These models often do not assume a parametric form for the decoder, enhancing their flexibility. For instance, \citet{xiao2017modeling} propose the use of W-GANs to generate new events. Similarly, \citet{shchurintensity} learn a distribution over the next event time via a normalizing flow. %
\citet{lin2022exploring} benchmark several choices of generative models, including diffusion, GANs, and VAEs. Regardless of the specific choice of decoder, these approaches are all autoregressive, making them ill-suited for multi-step forecasting tasks. Notable exceptions are the works of \citet{ludke2023add, ludke2026editbased}, which propose flow and diffusion based models that avoids autoregressive sampling through an iterative refinement procedure.%

\looseness=-1
Our \texttt{EventFlow} model can be viewed as a flexible, non-autoregressive decoder for TPPs, obtained by extending flow matching to continuous-time event sequences. Among existing approaches, \citet{ludke2023add, ludke2026editbased} are most closely related in spirit. However, their method relies on a relatively involved training and sampling pipeline based on iterative refinement. In contrast, \texttt{EventFlow} admits a particularly simple regression-based formulation which is straightforward to implement. %
While previous works have studied marked TPPs \citep{du2016recurrent, mei2017neural, draxler2025transformers, changdeep}, \texttt{EventFlow} is focused on modeling the event times themselves.

\looseness=-1
\paragraph{Flow Matching} The flow matching framework (or stochastic interpolants) \citep{lipman2023flow, lipman2024flow, albergo2022building, liu2022flow} describes a class of generative models which are closely related to normalizing flows \citep{papamakarios2021normalizing}. %
Flow matching has been explored for 
applications including image generation \citep{dao2023flow,ma2024sit}, DNA and protein design \citep{stark2024dirichlet, campbell2024generative}, %
and 3D modeling \citep{buhmann2023epic, wu2023fast, xiang2025structured}, but our work is the first to apply flow matching for TPPs.

\section{Autoregressive TPP Models}
\label{sec:background}
\looseness=-1
We first provide a brief, informal review of autoregressive point process models and discuss their shortcomings. Informally, an event sequence is a set $\{ t^k\}_{k=1}^n$ of increasing event times. We will use $\HH_{t}$ to represent the history of a sample up to (and including) time $t$, i.e., $\HH_t = \{ t^k : t^k \leq t \}$. Similarly, $\HH_{t^{-}} = \{ t^k : t^k < t \}$ represents the history prior to time $t$. In the autoregressive setting, the time of a single future event $t$ is modeled conditioned on the observed history of a sequence. This is often achieved by either directly modeling a distribution over $t$ \citep{shchurintensity}, or equivalently by modeling a conditional intensity function \citep{du2016recurrent}.

More precisely, one models conditional densities of the form $p(t^{k} \mid \HH_{t^{k-1}})$, which describe the distribution of the next event time given the history up to the previous event. This induces a joint distribution over event times $p(t^1, \dots, t^n)$ autoregressively via ${p(t^1, \dots, t^n) = p(t^1) \prod_{k=2}^n p(t^{k} \mid \HH_{t^{k-1}})}$.
Alternatively, we may define the \textit{conditional intensity} $\lambda^*(t) := \lambda(t \mid \HH_{t^-}) = p(t \mid \HH_{t^n})/(1 - F(t \mid \HH_{t^n}))$,
where $F(t \mid \HH_{t^n}) = \int_{t^n}^t p(s \mid \HH_{t^n}) \d s$ is the CDF associated with $p(t \mid \HH_{t^n})$. Informally, the conditional intensity $\lambda^*(t)$ can be thought of \citep{rasmussen2011temporal} as the instantaneous rate of occurrence of events at time $t$ given the history up to $t^n$ and that no events have occurred in $(t^n, t)$. 
By integrating $\lambda^*(t)$, one can verify
\begin{equation} \label{eqn:density_to_intensity_cdf}
    F(t \mid \HH_{t^n}) = 1 - \exp\left(- \int_{t_n}^t \lambda^*(s) \d s \right) 
\end{equation}
\begin{equation}  \label{eqn:density_to_intensity}
    p(t \mid \HH_{t^n}) = \lambda^*(t) \exp\left( - \int_{t_n}^t \lambda^*(s) \d s \right)
\end{equation}

and thus one may recover the conditional distribution from the conditional intensity under mild additional assumptions \citep[Prop~2.2]{rasmussen2011temporal}.%

\paragraph{The Likelihood Function}

Suppose we observe an event sequence $\{ t^k \}_{k=1}^n$ on the interval $[0, T]$. The \textit{likelihood} of this sequence can be seen loosely as the probability of observing events at these times and no others within the observation window. The likelihood may be expressed in terms of either the density or intensity via
\begin{align} \label{eqn:likelihood}
    L\left(\{ t^k\}_{k=1}^n\right) &= p(t^1, \dots, t^n) \left( 1 - F(T \mid \HH_{t^n}) \right) \\
    &= \left( \prod_{k=1}^n \lambda^*(t^k) \right) \exp\left(-\Lambda^*(T)\right)
\end{align}
where $\Lambda^*(T) = \int_0^T \lambda^*(s) \d s$ is the total intensity. The survival term $1 - F(T \mid \HH_{t^n})$ accounts for the absence of events in $(t^n, T]$. Autoregressive models are typically trained by maximizing this likelihood \citep{du2016recurrent, mei2017neural, shchurintensity}. Critically, we emphasize that $p(t^1, \dots, t^n)$ captures a joint density over $n$ event times, but does not account for what occurs after the final event $t^n$. In contrast, the likelihood $L\left(\{ t^k\}_{k=1}^n\right)$ corresponds to observations in a finite window $[0, T]$, and therefore includes an additional CDF term to account for the fact that no events were observed in $(t^n, T]$.

It is worth noting that evaluating $L(\{ t_k \})$ can be non-trivial in practice. For models that parametrize $\lambda^*(t)$ via a neural network \citep{du2016recurrent, mei2017neural}, computing the total intensity $\Lambda^*(T)$ is often done via a Monte Carlo integral, requiring many forward passes of the model to evaluate $\lambda^*(t)$ at different values of $t$. Models which directly parametrize the density $p(t \mid \HH_t)$ suffer from the same drawback when computing the corresponding CDF in Equation \eqref{eqn:likelihood}. Moreover, some approaches, such as the diffusion-based approach of \citet{lin2022exploring}, are only trained to maximize an ELBO of $p(t_k \mid t_1, \dots, t_{k-1})$, and are thus unable to compute the proper likelihood in Equation \eqref{eqn:likelihood}.

\paragraph{Sampling Autoregressive Models}
In many tasks, we are interested not only in an accurate model of the intensity (or distribution), but also sampling new event sequences. For instance, when forecasting an event sequence, we may want to generate several forecasts in order to provide uncertainty quantification over these predictions. However, sampling from existing models can be difficult. The flow-based model of \citet{shchurintensity} requires a numerical approximation to the inverse of the model to perform sampling. The diffusion-based approach of \citet{lin2022exploring} can require several hundred forward passes of the model to generate a \textit{single} event time, rendering it costly when generating long sequences. Moreover, the predictive performance of autoregressive models is often unsatisfactory on multi-step generation tasks due to the accumulation of errors over many steps \citep{lin2021empirical, ludke2023add}. This difficulty is particularly pronounced for intensity-based models \citep{du2016recurrent, mei2017neural, zhang2020self}, where naively computing the implied distribution in Equation \eqref{eqn:density_to_intensity} is prohibitively expensive. Instead, sampling from intensity-based models is typically achieved via the thinning algorithm \citep{ogata1981lewis, lewis1979simulation}. However, this algorithm has several hyperparameters to tune, is challenging to parallelize, and can be difficult for practitioners to implement \citep{xue2023easytpp}. For instance, the thinning algorithm requires one to know an upper bound on the intensity, which is in practice selected heuristically.

\section{EventFlow}
\label{sec:method}

\looseness=-1
Motivated by the limitations of autoregressive models, we propose \texttt{EventFlow}, which has a number of distinct advantages over prior work. First, \texttt{EventFlow} directly models the joint distribution over event times, thereby avoiding autoregression entirely. Second, our model is likelihood-free at training time, avoiding the Monte Carlo estimates needed to estimate the likelihood in Equation \eqref{eqn:likelihood}. Third, sampling from our model amounts to solving an ordinary differential equation. This is straightforward to implement and parallelize, allowing us to avoid the difficulties of thinning-based approaches. In fact, \texttt{EventFlow} can achieve competitive performance using only a \textit{single} forward pass at sampling time. 

We build upon the flow matching  (or stochastic interpolant) framework \citep{lipman2023flow, albergo2022building, liu2022flow} to develop our model. We begin below by focusing on the unconditional setting (i.e., generating an event sequence without being given an observed history), followed by an extension for conditional generation.

\paragraph{Preliminaries}

We first introduce some necessary background and notation. Let $\TT = [0, T]$ be a finite length interval. The set $\Gamma$ denotes the \textit{configuration space} of $\TT$ \citep{albeverio1998analysis}, i.e., the set of all finite counting measures on the set $[0, T]$. A point $\gamma \in \Gamma$ corresponds to a measure of the form $\gamma = \sum_{k=1}^n \delta[t^k]$, i.e., a finite collection of Dirac deltas located at event times $t^k \in \TT$. A \emph{temporal point process (TPP)} on $\TT$ is a probability distribution $\mu$ over the configuration space $\Gamma$. We use $\P(\Gamma)$ to denote the set of all such distributions. Informally, a TPP $\mu$ is a distribution over finite sequences $\gamma \in \Gamma$ whose events are in $\TT$. We use $N: \Gamma \to \Z_{\geq 0}$ to denote the counting functional, i.e., $N(\gamma)$ is the number of events in the sequence $\gamma$.\footnote{This can be thought of in terms of the counting process, i.e., $N(\gamma)$ corresponds to the value of the associated counting process at the ending time $T$, or the total number of events in $\gamma$ that have occurred in the interval $[0, T]$.} While it is common to represent TPPs as distributions over random sets of event times, in our approach it will be more convenient to represent TPPs as random measures \citep{kallenberg2017random}.

Under mild assumptions, a TPP $\mu$ can be fully characterized \citep[Prop.~5.3.II]{daley2003introduction} by a probability distribution which specifies the number of events and a \textit{collection} of joint densities corresponding to the event times themselves. In a slight abuse of notation, we will write $\mu(n)$ for the corresponding distribution over event counts, and $\{\mu^n(t^1, \dots, t^n) \}_{n=1}^\infty$ for the collection of joint distributions. In other words, for any given $n \in \Z_{\geq 0}$, the probability of observing $n$ events in the interval $\TT$ is $\mu(n)$, and $\mu^n(t^1, \dots, t^n)$ describes the corresponding joint distribution of event times. We further restrict each $\mu^n$ to be supported only on the ordered sets, so that $t^1 < t^2 < \dots < t^n$.  

Let $\mu_1$ represent the data distribution and $\mu_0$ represent a reference distribution, i.e., both $\mu_0, \mu_1 \in \P(\Gamma)$ are distributions over event sequences. To construct our model, we will define a \textit{path} of distributions $(\eta_s)_{s \in [0.1]} \subset \P(\Gamma)$ which approximately interpolates from our reference TPP to our data TPP. Intuitively, this path provides us with a way to \textit{transform} samples: we can draw a sequence of events from $\mu_0$ and gradually move them along this path to obtain a sequence approximately distributed according to $\mu_1$. Throughout, we use $s \in [0, 1]$ to denote a flow time and $t \in [0, T]$ to denote an event time. These two time axes are in a sense orthogonal to one another (see Figure \ref{fig:leading_figure}). 

\looseness=-1
\paragraph{Balanced Couplings} Towards constructing such a path, our first step is to define a notion of \textit{coupling}, which allows us to pair event sequences drawn from the reference distribution $\mu_0$ with those drawn from the data distribution $\mu_1$. Formally, a \textit{coupling} between two TPPs $\mu_0, \mu_1\in \P(\Gamma)$ is a joint probability distribution $\rho \in \P(\Gamma \times \Gamma)$ over pairs of event sequences $(\gamma_0, \gamma_1)$  such that the marginal distributions of $\rho$ are $\mu_0$ and $\mu_1$. Sampling $(\gamma_0, \gamma_1) \sim \rho$ from a coupling gives us paired sequences of events.

While couplings are broadly used in transport-based generative modeling \citep{tong2023improving, villani2009optimal}, we introduce a new class of couplings which is particularly well-suited to the TPP setting. We say that the coupling $\rho$ is \textit{balanced} if draws $(\gamma_0, \gamma_1) \sim \rho$ are such that $N(\gamma_0) = N(\gamma_1)$ almost surely. That is, balanced couplings only pair event sequences with equal numbers of events. This ensures that when we later construct interpolations between paired sequences, we do not need to add or remove events, simplifying both training and sampling. 
We will use $\Pi_b(\mu_0, \mu_1)$ to denote the set of balanced couplings. 

The following proposition characterizes when balanced couplings exist. In particular, $\Pi_b(\mu_0, \mu_1)$ is nonempty if and only if the event count distributions of $\mu_0$ and $\mu_1$ are identical, imposing a structural constraint on suitable choices of $\mu_0$: the reference TPP $\mu_0$ must have the same event count distribution as the data TPP $\mu_1$.

\begin{restatable}[Existence of Balanced Couplings]{prop}{balancedcouplings} \label{prop:balanced_couplings}
    Let $\mu_0, \mu_1 \in \P(\Gamma)$ be two TPPs. The set of balanced couplings $\Pi_b(\mu_0, \mu_1)$ is nonempty if and only if have the same distribution over event counts, i.e., $\mu_0(n) = \mu_1(n)$ for every $n \in \Z_{\geq 0}$.
\end{restatable}
We provide a proof in Appendix \ref{appendix:proofs}. %

In practice, we follow a simple strategy for choosing both the reference TPP $\mu_0$ and the coupling $\rho$. Let $q$ be a density on $\TT$ (e.g., uniform). We take $\mu_0$ to be a mixed binomial process \citep[Ch.~3]{kallenberg2017random} whose event count distribution is given by that of the data $\mu_1(n)$, and joint event time distributions given by independent products of $q$ (up to sorting). Sampling $\gamma_0 \sim \mu_0$ amounts to sampling $n \sim \mu_1(n)$ from the data event count distribution followed by sampling and sorting $n$ i.i.d. times $t^k \sim q$. Under this choice, a sample from a balanced coupling $\rho \in \Pi_b(\mu_0, \mu_1)$ can be produced by first sampling a data sequence $\gamma_1 \sim \mu_1$, followed by sampling $N(\gamma_1)$ events independently from $q$ and sorting to produce a paired draw $\gamma_0 \sim \mu_0$. %

\paragraph{Interpolant Construction}
We now construct our path $(\eta_s)_{s \in [0, 1]} \subset \P(\Gamma)$ using a two-stage procedure. First, for a given pair of sequences $(\gamma_0, \gamma_1)$ drawn from a balanced coupling, we define a sequence-level interpolation $(\gamma_s)_{s\in[0,1]}$ that smoothly transforms $\gamma_0$ into $\gamma_1$. Second, the distribution $\eta_s$ is obtained by averaging these interpolated sequences over the coupling. We extend flow matching techniques \citep{lipman2023flow, tong2023improving} to define the interpolants, and emphasize that fixing the number of events via a balanced coupling is essential for this construction.

To that end, let $\rho$ be any balanced coupling of the reference distribution $\mu_0$ and the data distribution $\mu_1$, and suppose $z := (\gamma_0, \gamma_1) \sim \rho$ is a draw from this coupling. As $\rho$ is balanced, we have $\gamma_0 = \sum_{k=1}^n \delta[t_0^k]$ and $\gamma_1 = \sum_{k=1}^n \delta[t_1^k]$
are both a collection of $n$ events. We will henceforth describe our procedure for a fixed (but arbitrary) number of events $n$, and we will later describe how to model the number of events itself. First, we define the interpolant sequence $\gamma_s^z \in \Gamma$ via
\begin{equation} \label{eqn:deterministic_path}
    \gamma_s^z = \sum_{k=1}^n \delta[t_s^k] \qquad t_s^k = (1-s)t_0^k + s t_1^k \qquad 0 \leq s \leq 1
\end{equation}
where we use the superscript $z$ to denote the dependence on the pair $z = (\gamma_0, \gamma_1)$. In other words, $\gamma_s^z$ linearly interpolates each corresponding event in $\gamma_0$ and $\gamma_1$, %
defining a path $(\gamma_s^z)_{s\in[0,1]} \subset \Gamma$ which evolves the reference sample $\gamma_0$ into the data sample $\gamma_1$. 

\looseness=-1
We now lift this deterministic path $(\gamma_s^z)_{s\in[0,1]} \subset \Gamma$ to a path of TPP distributions $(\eta_s^z)_{s\in[0, 1]} \subset \P(\Gamma)$. We define the point process distribution $\eta_s^z \in \P(\Gamma)$ implicitly by adding independent Gaussian noise to each event in $\gamma_s^z$. That is, a draw $\hat{\gamma}_s^z \sim \eta_s^z$ may be simulated via
\begin{equation} \label{eqn:conditional_path_measures}
    \hat{\gamma}_s^z = \sum_{k=1}^n \delta\left[t_s^k + \epsilon^k\right] \qquad \epsilon^k \sim \NN(0, \sigma^2).
\end{equation}
In principle, using Gaussian noise means that the support of event times in $\eta_s^z$ is larger than $\TT$, but in practice we choose $\sigma^2$ sufficiently small such that this is not a concern. The addition of noise $\epsilon^k$ is instrumental in obtaining a well-specified model, but in practice we found the noise variance $\sigma^2$ to not be a critical hyperparameter. We note that this noising step is typical in flow matching models \citep{lipman2023flow, tong2023improving}.

\looseness=-1
Finally, for $s \in [0, 1]$, we define the marginal TPP measure $\eta_s \in \P(\Gamma)$ via the mixture distribution
\begin{equation} \label{eqn:marginal_measure}
    \eta_s = \int \eta_s^z \d \rho(z)
\end{equation}
which, intuitively, corresponds to the two-stage procedure of drawing a sample $z \sim \rho$ and subsequently following the conditional flow defined in Equation~\eqref{eqn:conditional_path_measures}.

By construction, the event count distribution $\eta_s(n)$ is given by $\mu_1(n)$ for all $s \in [0, 1]$. This path of TPP distributions $\eta_s$ approximately interpolates from the reference TPP $\mu_0$ at $s=0$ to the data TPP $\mu_1$ at $s=1$, in the sense that at the endpoints, the joint event time distributions $\eta_0^n(t^1, \dots, t^n)$ and $\eta_1^n(t^1, \dots, t^n)$ are given by a convolution of $\mu_0^n(t^1, \dots, t^n)$ and $\mu_1^n(t^1, \dots, t^n)$ with the Gaussian $\NN(0, \sigma^2 I_n)$. As $\sigma^2 \downarrow 0$, it is clear that we recover a genuine interpolant.

Observe that in Equation~\eqref{eqn:conditional_path_measures}, for each individual event $t_s^k$, we have constructed a path of Gaussian distributions $\NN(t_s^k, \sigma^2)$ centered around said event. This path of Gaussians can be simulated via the vector field $v_s^k := t_1^k - t_0^k$ \citep{tong2023improving}. That is, if we draw an initial condition $\tau_0^k \sim \NN(t_0^k, \sigma^2)$ and solve the differential equation $\d \tau_s^k = v_s^k \d s$ for $s \in [0, 1]$ starting from $\tau_0^k$, we obtain a draw $\tau_1^k \sim \NN(t_1^k, \sigma^2)$. Loosely, $v_s^k$ is a velocity associated with the event $t_s^k$.

Since each event in Equation~\eqref{eqn:conditional_path_measures} evolves independently, the path of distributions ${\eta}_s^z$
is generated by the constant vector field $v_s^z: \TT^n \to \R^n$ defined as
\begin{equation}
    v_s^z(\gamma) := \begin{bmatrix} v_s^1, \dots, v_s^n \end{bmatrix}^{\sf T} =\begin{bmatrix} t_1^1 - t_0^1, & \dots , & t_1^n - t_0^n \end{bmatrix}^{\sf T}.%
\end{equation}
In other words, flowing initial samples $\hat{\gamma}_0^z \sim \eta_0^z$ along the vector field $v_s^z(\gamma)$ generates the path of distributions $\eta_s^z$, approximately interpolating from the noise sequence $\gamma_0$ to the data sequence $\gamma_1$.

However, this path is conditioned on $z = (\gamma_0, \gamma_1)$. Since we seek to generate $\gamma_1$, this is intractable, and we would instead like to find the vector field $v_s$ that generates the \textit{unconditional} path $\eta_s$ defined in Equation~\eqref{eqn:conditional_path_measures}. 
To this end, we aggregate the conditional vector fields $v_s^z(\gamma)$ across the coupling $z \sim \rho$. In particular, the 
unconditional vector field $v_s: \TT^n \to \R^n$ is given by
\begin{equation} \label{eqn:marginal_vf}
    v_s(\gamma) = \E_{z \sim \rho}[v_s^z(\gamma) \mid \gamma_s^z = \gamma] = \int v_s^z(\gamma) \frac{\d \eta_s^z}{\d \eta_s} (\gamma) \d \rho(z)
\end{equation}
i.e., the average of the conditional vector fields over all pairs $z$ that could have produced the sequence $\gamma$ at time $s$. This corresponds to an extension of the standard flow matching construction \citep{lipman2023flow, tong2023improving, albergo2022building} to the setting of TPPs.

\begin{algorithm}[t]
\DontPrintSemicolon
\caption{Training \texttt{EventFlow}} \label{alg:training}
Sample $\gamma_1 \sim \mu_1$, $s \sim \UU[0, 1]$, $\epsilon \sim \NN(0, 1)$ \;
$e_{\HH} = \varnothing$ \Comment*[r]{Null history}
\If{forecast}{
  Sample $T_0 \in [\Delta T, T - \Delta T]$\;
  $\HH \gets \{ t \in \gamma_1 : t \leq T_0 \}$\;
  $e_{\HH} \gets f_\theta(\HH)$\;
  $\gamma_1 \gets \{ t \in \gamma_1 : T_0 < t \leq T_0 + \Delta T \}$\;
}
$n \gets N(\gamma_1)$\;
Sample and sort $t_0^k \sim q$ for $k = 1, \dots, n$\;
$\gamma_0  \gets (t_0^1, \dots, t_0^n)$\;
$t_s^k \gets (1-s) t_0^k + s t_1^k$ for $k = 1, \dots, n$ \;
$\gamma_s^z \gets (t_s^1, \dots, t_s^n)$ \;
Take a gradient step on $\norm{\gamma_1 - \gamma_0 - v_\theta\left(\gamma_s^z + \epsilon, s, e_{\HH} \right)}^2$
\end{algorithm}

\paragraph{Training} If we knew $v_s$ in Equation \eqref{eqn:marginal_vf}, we could draw a sample from $\gamma_1 \sim \mu_1$ by drawing a sample event sequence $\gamma_0 \sim \mu_0$ from the reference TPP and flowing each event along $v_s$. More precisely, $\gamma_0$ will consist of $N(\gamma_0)$ events and $v_s(\gamma)$ will be a vector field with $N(\gamma_0)$ components, so that we may solve the differential equation $\d \gamma_s = v_s(\gamma) \d s$ on $s \in [0, 1]$. 

Although the marginal vector field in Equation \eqref{eqn:marginal_vf} admits an analytical form, it is intractable to compute in practice as the density ratio $\d \eta_s^z/\d \eta_s$ is unknown. To overcome this, we instead regress on the \textit{conditional} vector fields $v_s^z$. Here, $v_\theta(\gamma_s, s)$ will represent a neural network with parameters $\theta$ which takes in a sequence $\gamma_s$ along with the flow time $s$. We aim to minimize
\begin{equation} \label{eqn:loss}
    J(\theta) = \E_{s, (\gamma_0, \gamma_1), \hat{\gamma}_s^z} \left[\norm{\gamma_1 - \gamma_0 - v_\theta(\hat{\gamma}_s^z, s)}^2 \right]
\end{equation}
which is equal to the MSE loss on the \textit{unconditional} $v_s$ up to an additive constant not depending on $\theta$ \citep{lipman2023flow, tong2023improving}. %
We estimate $J(\theta)$ by uniformly sampling a flow time $s \in [0, 1]$, a pair $z = (\gamma_0, \gamma_1) \sim \rho$ from our balanced coupling and drawing a noisy interpolant $\hat{\gamma}_s^z \sim \eta_s^z$ according to Equation \eqref{eqn:conditional_path_measures}.

\looseness=-1
To train the model for forecasting, where the goal is to predict a future sequence of events conditioned on a history $\HH$, we embed $\HH$ into a fixed-dimensional vector representation $e_\HH = f_\theta(\HH)$ via a learned encoder $f_\theta$ before providing this to the model $v_\theta(\gamma_s, s, e_{\HH})$ and minimizing Equation \eqref{eqn:loss}. Note that we jointly train the encoder $f_\theta$ and vector field $v_\theta$. See Algorithm \ref{alg:training} for training pseudocode.

\paragraph{Event Count Distributions} We have thus far described a procedure for interpolating between a given reference distribution $\mu_0^n$ and the data distribution $\mu_1^n$ for a given, fixed number of events $n$. As $n$ was arbitrary, we have successfully constructed a family of interpolants which will enable us to sample from the joint event time distributions $\mu_1^n(t^1, \dots, t^n)$. However, recall that fully characterizing the TPP distribution requires us to also specify the event \textit{count} distribution. 

For unconditional generation tasks, this is straightforward: we simply follow the empirical event count distribution see in the training data. When forecasting, though, we must also learn a model of the event count distribution $p_\phi(n \mid \HH)$, i.e., how many events are likely to occur after observing a history $\HH$. Here, $\phi$ represents the parameters of this model. We train $p_\phi(n \mid \HH)$ by minimizing a cross-entropy loss. In practice, we found it important to regularize this loss to encourage smoothness in $n$, which we achieve through an optimal-transport based regularizer. See Appendix \ref{appendix:model_arch} for details.

As both $v_\theta$ and $p_\phi$ must be able to operate on variable-length sequences, we use a transformer-based backbone for both models. However, our overall method is agnostic to this choice.

\begin{algorithm}[t]
\DontPrintSemicolon
\caption{Sampling \texttt{EventFlow}} \label{alg:sampling}
Choose a flow time discretization $0 = s_0 < s_1 < \dots < s_K = 1$\;
$e_{\HH} = \varnothing$ \Comment*[r]{Null history}
\eIf{forecast}{
  $e_\HH \gets f_\theta(\HH)$\;
  $n \sim p_\phi(n \mid \HH)$\;
}{
  $n \sim \mu_1(n)$\;
}
Sample and sort $t_0^k \sim q$ for $k = 1, \dots, n$\;
$\gamma_0  \gets (t_0^1, \dots, t_0^n)$\;
\For{$k = 1, 2, \dots, K$}{
  $h_k \gets s_k - s_{k-1}$\;
  $\gamma_{s_k} \gets \gamma_{s_{k-1}} + h_k v_\theta(\gamma_{s_{k-1}}, s_{k-1}, e_{\HH})$\;
}
\Return $\gamma_0$
\end{algorithm}

\paragraph{Sampling}
Once $v_\theta$ is learned, we may sample from the model by drawing a reference sequence $\gamma_0 \sim \mu_0$ and solving the corresponding ODE parametrized by $v_\theta$. We fix a number of events $n$, which can be sampled from the empirical event count distribution $\mu_1(n)$ for unconditional generation.
For conditional tasks, we draw $n \sim p_\phi(n \mid \HH)$ from the learned conditional distribution over event counts. Next, we draw $n$ initial events, corresponding to $s = 0$, by sampling and sorting $t_0^1, \dots, t_0^n \sim q$. In practice, we use $q = \NN(0, I_n)$ as we normalize our sequences into the range $[-1, 1]$ during training and sampling (followed by renormalization to the data scale). Since we have fixed $n$, we may view this initial draw as a vector $\gamma_0 = [t_0^1, \dots, t_0^n] \in \TT^n$. This event sequence $\gamma_0$ then serves as the initial condition for the system of ODEs $d \gamma_s = v_\theta(\gamma_s, s) \d s$ which can be solved numerically. In our experiments, we use the forward Euler scheme, i.e., we specify a discretization $\{0 = s_0 < s_1 < \dots < s_K = 1\}$ of the flow time (in practice, uniform) and recursively compute
\begin{equation} \label{eqn:diffeq}
    \gamma_{s_k} = \gamma_{s_{k-1}} + h_k v_\theta(\gamma_{s_{k-1}}, s_{k-1}), \quad k = 1, \dots, K
\end{equation}
where $h_k = s_k - s_{k-1}$ is a step size. While other choices are possible, we found that this simple scheme was sufficient as the model sample paths are, qualitatively, typically close to linear. See Algorithm \ref{alg:sampling}.

\section{Experiments}
\label{sec:experiments}

\begin{figure*}[t]
    \centering
    \includegraphics[width=5.52in]{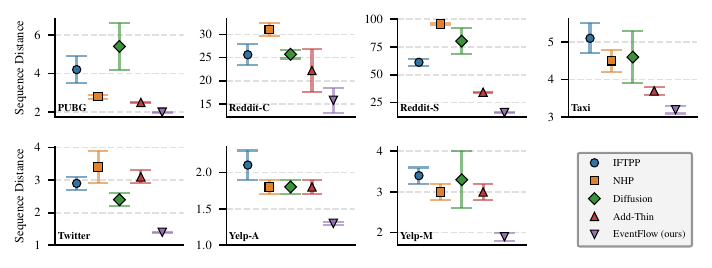}
    \caption{Sequence distance \eqref{eqn:sequence_distance} between the forecasted and ground-truth event sequences on a held-out test set. We report the mean $\pm$ one standard deviation over five random seeds. \texttt{EventFlow} (with 25 NFEs) achieves the lowest mean distance (forecasting error) for each of the 7 datasets.}
    \label{fig:sequence_distance_results}
\end{figure*}

We study our proposed \texttt{EventFlow} model under two settings\footnote{Code for our experiments is available at this URL: \url{https://github.com/GavinKerrigan/eventflow}}. The first is a conditional task, where we seek to forecast both the number of future events and their times of occurrence, over a particular horizon, given a history. The second is an unconditional task, where we aim to learn a representation of the underlying TPP distribution from empirical observations and generate new sequences from this distribution. This second task can be viewed as a special case of the first with no observed history.

We evaluate our model across a diverse set of datasets encompassing a wide range of possible point process behaviors. We  evaluate our model on seven real-world datasets, which are a standard benchmark for modeling unmarked TPPs \citep{shchur2020fast, bosser2023predictive, ludke2023add}. We also evaluate on a collection of six synthetic datasets \citep{omi2019fully}. See Appendix \ref{appendix:datasets} for additional information regarding our datasets.

\paragraph{Baseline Models} We selected a set of baselines consisting of a set of diverse and highly performant neural TPP models. For an intensity-based method, we compare against the Neural Hawkes Process (NHP) \citep{mei2017neural}, a well-known and very widely-used neural TPP model. We additionally compare against two intensity-free methods, namely the flow-based IFTPP model \citep{shchurintensity} and the diffusion-based model of \citet{lin2022exploring}. Lastly, our strongest baseline is the Add-and-Thin model of \citet{ludke2023add}, which can be loosely viewed as a non-autoregressive diffusion model. These models use an RNN-based history encoder, with the exception of Add-and-Thin which uses a CNN-based encoder. See Appendix \ref{appendix:baselines} for additional details on our baselines.

\paragraph{Metrics}
Evaluating generative TPP models is challenging, as one must take into account both the variable locations and numbers of events. This is particularly challenging for the unconditional setting, where unlike forecasting, we do not have a ground-truth sequence.%

Our starting point is a metric \citep{xiao2017wasserstein} on the space of sequences $\Gamma$, allowing us to measure the distance between two sequences $\gamma = \sum_{k=1}^n \delta[t^k_\gamma]$ and $\eta = \sum_{k=1}^m \delta[t^k_\eta]$ with possibly different numbers of events. Without loss of generality, we assume $n \leq  m$, in which case the distance is given by
\begin{equation} \label{eqn:sequence_distance}
    d(\gamma, \eta) = \sum_{k=1}^n |t^k_\gamma - t^k_\eta| + \sum_{k=n+1}^m (T - t^k_{\eta})
\end{equation}
where we recall that sequences are supported on $\TT = [0, T]$. This distance can be understood either as an $L^1$ distance between the  counting processes of $\gamma, \eta$ or as a generalization of the $1$-Wasserstein distance to measures of unequal mass. For our unconditional experiment, we require a metric that will capture the distance between the TPP distributions themselves. To do so we use the distance in Equation \eqref{eqn:sequence_distance} to calculate an MMD \citep{gretton2012kernel, shchur2020fast}. 
We use an exponential kernel $k(\gamma, \eta) = \exp\left( - d(\gamma, \eta) / (2\sigma^2)  \right)$ with $\sigma$ chosen to be the median distance between all sequences in a dataset \citep{shchur2020fast, ludke2023add}.

\subsection{Forecasting Event Sequences}
\label{section:forecasting}

\looseness=-1
We first evaluate our model on a multi-step forecasting task. We set a horizon $\Delta T$ for each of our real-world datasets and seek to generate event sequences in the range $[T_0, T_0 + \Delta T]$ for some given $T_0$, conditioned on the history $\HH_{T_0}$. Up to a shift, this means we are taking $\TT = [0, \Delta T]$. %
At training time, we uniformly sample $T_0 \in [\Delta T, T - \Delta T]$. At test time, we sample 50 values of $T_0$ for each test set sequence. We then generate one forecast for the sequence in $[T_0, T_0 + \Delta T]$ and compute the distance \eqref{eqn:sequence_distance} between the ground-truth and generated sequences. We note that the distance in Equation \eqref{eqn:sequence_distance} is computed using $T_0 + \Delta T$ rather than $T$ as the maximum event time, as using $T$ would result in a distance which is sensitive to the location of the forecasting window. We further normalize Equation \eqref{eqn:sequence_distance} by $\Delta T$. This overall approach is similar to that of \citet{ludke2023add}. For all datasets, the sequences are split into train-validation-test sets, and all metrics are reported on the held-out test set.

We report the results of these experiments in Figure \ref{fig:sequence_distance_results}. %
Our proposed \texttt{EventFlow} method obtains the lowest average forecasting error across all datasets, achieving an error which is 20\%-53\% lower than the nearest baseline. Given that the non-autoregressive models (\texttt{EventFlow}, Add-and-Thin) typically outperform the autoregressive baselines, we see clear evidence that autoregressive models can struggle on multi-step tasks. This is especially true on Reddit-C and Reddit-S which exhibit long sequence lengths. %

\begin{table}[!t]
\caption{Mean sequence distance for \texttt{EventFlow} as we vary the NFEs used to simulate the ODE.}
\vspace{0.5em}
\label{tab:forecast_ablation}
\centering
\begingroup
\setlength{\tabcolsep}{2.8pt}
\small{
\begin{tabular}{@{}lrrrrrrr@{}}
\toprule
NFEs & PUBG & Red.C & Red.S & Taxi & Tw. & Yelp-A & Yelp-M \\ \midrule
25 & $ 2.0 $ & $15.8 $& $16.0$ & ${3.2}$ & ${1.4}$ & ${1.3} $ & ${1.9}$ \\
10  & $2.0 $& $15.8$& $15.8$& $3.1$ & $1.4$ & $1.3$ & $1.9$ \\
1 & $2.0$& $15.8$& $15.8$& $3.7$& $1.4$ & $1.8$& $1.9$ \\
\bottomrule
\end{tabular}
}
\endgroup
\end{table}

\begin{table}[!h]
\caption{MARE values evaluating the predicted number of events when forecasting. Mean values are reported over five random seeds. %
}
\vspace{0.5em}
\label{tab:mape}
\centering
\begingroup
\setlength{\tabcolsep}{1.7pt}
\renewcommand{\arraystretch}{1.1}
\small{
\begin{tabular}{@{}l@{\hspace{-3pt}}rrrrrrr@{}}
\toprule
                            & PUBG & Red.C & Red.S & Taxi & Tw. & Yelp-A & Yelp-M \\ \midrule
                        IFTPP & $1.05$ & $1.69$ & $0.79$ & $0.60$ & $0.88$ & $0.76$ & $0.76$ \\ 
                        NHP   & $1.02$ & $\underline{0.95}$ & $1.00$ & $0.67$ & $2.48$ & $0.80$ & $1.07$ \\
                    Diffusion & $1.95$  & $1.28$ & $1.12$ & $0.49$ & $0.66$ & $0.65$ & $0.72 $\\
               Add-Thin & $\underline{0.43}$ & $0.99$ & $\underline{0.38}$ & $\underline{0.33}$ & $\underline{0.60}$ & $\mathbf{0.42}$ &  $\mathbf{0.46}$\\
                       EventFlow & $\mathbf{0.40}$ &  $\mathbf{0.70}$ & $\mathbf{0.16}$& $\mathbf{0.28}$ & $\mathbf{0.46}$ & $ \underline{0.56}$ & $\underline{0.50}$ \\ \bottomrule
\end{tabular}
}
\endgroup
\end{table}

\looseness=-1
In Table \ref{tab:forecast_ablation}, we additionally vary the number of function evaluations (NFEs) used at test time, i.e., the number of steps $K$ used in the ODE solver in Algorithm \ref{alg:sampling}. We find that reducing $K$ results in minimal (or no) loss in performance. Even with only a single NFE, \texttt{EventFlow} is able to obtain state-of-the-art performance. This is due to the fact that, although the vector fields (Equation \eqref{eqn:marginal_vf}) in \texttt{EventFlow} are generally non-linear, we find qualitatively that our construction produces paths which are approximately so. In contrast, Add-and-Thin uses 100 NFEs, while the NFEs for the autoregressive models scale linearly with the number of generated events. %

\addtolength{\tabcolsep}{-0.3em}
\begin{table*}[!htb] 
\caption{MMDs (1e-2) between the test set and $1,000$ generated unconditional sequences averaged over five random seeds. \texttt{EventFlow} achieves the overall highest mean rank.}
\label{tab:mmd_all}
\centering
\vspace{0.1cm}
\small{
\begin{tabular}{lcccccc|ccccccc}
\toprule
                      & H1 & H2 & NSP   & NSR   & SC    & SR & PUBG & Red.C & Red.S & Taxi & Twitter & YelpA & YelpM \\ \midrule
IFTPP & $\mathbf{1.5}$ & $\mathbf{1.4}$ & $\mathbf{2.3}$ & $\underline{6.2}$ & $\mathbf{5.8}$ & $\mathbf{1.3}$ & $5.7$ & $1.3$ & $1.9$ & $5.8$ & $\mathbf{2.9}$ & $8.2$ & $\underline{5.1}$ \\
NHP & $\underline{1.9}$ & $5.2$ & $3.6$ & $12.6$ & $25.4$  & $5.0$ & $7.2$ & $2.2$ & $22.5$ & $\underline{5.0}$ & $7.3$ & $6.7$ & $6.1$ \\
Diffusion & $4.8$ & $5.5$ & $10.8$ & $15.0$ & $9.1$ & $5.1$ & $14.3$ & $3.9$ & $6.2$ & $11.7$ & $12.5$ & $10.9$ & $10.5$ \\
Add-Thin & $\underline{1.9}$ & $2.5$ & $\underline{2.6}$ & $7.4$ & $22.5$ & $2.2$ & $\underline{2.8}$ & $\underline{1.2}$ & $2.7$ & $5.2$ & $\underline{4.8}$ & $\mathbf{4.5}$ & $\mathbf{3.0}$ \\
EventFlow {\scriptsize (25 NFEs)} & $\underline{1.9}$ & $\underline{2.2}$ & $3.8$ & $\mathbf{4.2}$ & $\underline{8.3}$ & $\underline{1.7}$ & $\mathbf{1.5}$ & $\mathbf{0.7}$ & $\mathbf{0.7}$ & $\mathbf{3.5}$ & $4.9$ & $\underline{6.6}$ & $\mathbf{3.0}$\\
\bottomrule
\end{tabular}
}
\end{table*}

To evaluate the event count predictions, ${p_\phi(n \mid \HH)}$, we report the mean absolute relative error (MARE) between the true and predicted counts in Table \ref{tab:mape}. Our model achieves strong performance on this metric as it decouples the event count prediction from the event time prediction, training explicitly for each task. The baselines, on the other hand, only model $n$ implicitly. See Appendix \ref{appendix:additional_experiments} for details and standard deviations. %

\subsection{Unconditional Generation of Event Sequences}
Finally, we evaluate our model on an unconditional generation task, where we aim to generate new sequences from the underlying data distribution without conditioning on an observed history. %
In Table \ref{tab:mmd_all} we report MMD values for each of the synthetic and real-world datasets. %
MMDs are calculated by sampling 1,000 sequences from each trained model.
The MMD values in Table \ref{tab:mmd_all} correspond to the mean test set MMD ($\pm$ one standard deviation) across five random seeds. 
See Appendix \ref{appendix:additional_experiments} for results with standard deviations. 

\texttt{EventFlow} (mean rank: $\mathbf{1.8}$) exhibits uniformly strong performance, obtaining either the best or second best MMD on 11 of the 13 datasets. This is particularly pronounced on the real-world datasets, where we obtain the lowest MMD on 5 of the 7 datasets. 

Notably, IFTPP (mean rank: $\mathbf{2.1}$) is one of the strongest methods on the synthetic data, but is relatively weak on the real-world data. We believe that this can be understood in terms of model flexibility. IFTPP uses a simple parametric form (a mixture of log-normals) to parametrize the subsequent event time distribution (see Section~\ref{sec:background}). This provides a strong inductive bias and allows it to fit simple synthetic distributions very effectively, but limits its capability to model more realistic data sets. EventFlow, in contrast, makes no parametric assumptions, offering greater flexibility. As a result, it may not outperform simple parametric models on very simple synthetic datasets, but this flexibility is precisely what enables strong performance on complex real-world data, which we consider the more meaningful benchmark.

The Add-and-Thin method (mean rank: $\mathbf{2.4}$) is often similarly strong, but struggles on the SC dataset. While the NHP (mean rank: $\mathbf{3.7}$) can obtain good fits, this appears to be dataset dependent, with weak results on the NSR, SC, and Reddit-S datasets. The diffusion baseline (mean rank: $\mathbf{4.8}$) is our weakest baseline, which is perhaps unsurprising as this model can only be trained to maximize the likelihood of a subsequent event and not the overall sequence likelihood.%

\section{Conclusion}
\label{sec:conclusion}

We introduced \texttt{EventFlow}, a non-autoregressive generative model for temporal point processes. Our approach extends flow matching to the TPP setting by constructing a continuous transformation between a simple reference distribution and the data distribution over event sequences. Concretely, \texttt{EventFlow} generates event sequences by first drawing a sequence of event times from a reference process, and then transporting these events along a learned vector field. Beyond modeling event times, our framework also captures the distribution over the number of events, which we learn via a regularized cross-entropy objective.

Empirically, we demonstrate that \texttt{EventFlow} achieves state-of-the-art performance on multi-step forecasting tasks and competitive results on unconditional generation across a range of standard benchmarks. These results highlight the promise of non-autoregressive, flow-based approaches as a flexible alternative to classical intensity-based models for TPPs.

\paragraph{Limitations and Outlook} %
Our work focuses on modeling event times, and we do not consider marked or spatiotemporal point processes. Extending \texttt{EventFlow} to incorporate marks or spatial structure is a natural and promising direction for future work, though it may require new coupling and interpolation strategies.

Additionally, our current construction does not explicitly enforce support constraints on $[0, T]$, as it relies on Gaussian noising in Equation~\eqref{eqn:conditional_path_measures}. Addressing this limitation, e.g., by designing constrained flows, could potentially lead to further gains in performance. 
Finally, our experimental comparisons adopt the neural architectures used in prior work, which differ across baseline methods. A more controlled study of architectural choices, and their interaction with flow-based objectives, would help isolate the gains attributable to the modeling framework itself.

More broadly, we believe our work opens the door to a new class of generative models for point processes based on flow-based ideas, and we hope it encourages further exploration of non-autoregressive approaches.%

\FloatBarrier

\subsubsection*{Acknowledgements}
We thank the reviewers for their feedback on improving the paper. This work was supported  by National Science Foundation under awards  NSF 2505006 and NSF 2425932,  by the National Institutes of Health under awards  R01-LM013344 and R01CA297869, by the UK EPSRC grant EP/Y037200/1, by the Hasso Plattner Institute (HPI) Research Center in Machine Learning and Data Science at UCI, and by  funding support from Google and from SAP.

\bibliography{refs}
\bibliographystyle{plainnat}

\clearpage

\clearpage
\appendix
\thispagestyle{empty}

\onecolumn
\aistatstitle{EventFlow: Forecasting Temporal Point Processes with Flow Matching:\\ Supplementary Materials}

\section{Datasets}
\label{appendix:datasets}

In this section, we provide some additional details regarding the datasets used in this work. In Table \ref{tab:datasets}, we report the number of sequences in each dataset, some basic statistics regarding the number of events in each sequence, and their support $[0, T]$ and chosen forecast window $\Delta T$. In all datasets, we use 60\% of the sequences for training, 20\% for validation, and the remaining 20\% for testing. 

\paragraph{Synthetic Datasets} Our synthetic datasets are adopted from those proposed by \citet{omi2019fully}. Each of these datasets consists of $1,000$ sequences supported on $\TT = [0, 100]$. These synthetic datasets are chosen as they exhibit a wide range of behavior, ranging from i.i.d. inter-arrival times to self-correcting processes which discourage rapid bursts of events. We refer to Section 4 of \citet{omi2019fully} for details.

\paragraph{Real-World Datasets} We use the set of real-world datasets proposed in \citet{shchur2020fast}, which constitute a set of standard benchmark datasets for unmarked TPPs. We refer to Appendix D of \citet{shchur2020fast} for additional details. With the exception of PUBG, these datasets are supported on $\TT = [0, 24]$, i.e. each sequence corresponds to a single day. For the PUBG dataset, $\TT = [0, 38]$ corresponds to the maximum length (in minutes) of an online game of PUBG. We note that PUBG has the largest number of sequences (which can lead to slow training), and the Reddit-C and Reddit-S datasets have long sequences (which can lead to slow training and high memory costs). 

For selecting the forecasting horizon $\Delta T$, we follow the choice made in \citet{ludke2023add} to facilitate comparison and reproducibility. We note the initial times $T_0 \in [\Delta T, T - \Delta T]$ are selected to be at least $\Delta T$ to avoid forecasting sequences with no observations in the history.

\begin{table}[!hbt]
\caption{Some basic summary statistics of the datasets we consider in this work.}
\label{tab:datasets}
\centering
\small{
\begin{tabular}{@{}lcccccc@{}}
\toprule
                      & Sequences & Mean length & Std length & Range length & Support  & $\Delta T$      \\ \midrule
Hawkes1               & 1000      & 95.4        & 45.8       & [14, 300]    & [0, 100] & $\blank$        \\
Hawkes2               & 1000      & 97.2        & 49.1       & [18, 355]    & [0, 100] & $\blank$        \\
Nonstationary Poisson & 1000      & 100.3       & 9.8        & [71, 134]    & [0, 100] & $\blank$        \\
Nonstationary Renewal & 1000      & 98          & 2.9        & [86, 100]    & [0, 100] & $\blank$        \\
Stationary Renewal    & 1000      & 109.2       & 38.1       & [1, 219]     & [0, 100] & $\blank$        \\
Self-Correcting       & 1000      & 100.3       & 0.74       & [98, 102]    & [0, 100] & $\blank$        \\ \midrule
PUBG                  & 3001      & 76.5        & 8.8        & [26, 97]     & [0, 38]  & $5$             \\
Reddit-C              & 1356      & 295.7       & 317.9      & [1, 2137]    & [0, 24]  & $4$             \\
Reddit-S              & 1094      & 1129        & 359.5      & [363, 2658]  & [0, 24]  & $4$             \\
Taxi                  & 182       & 98.4        & 20         & [12, 140]    & [0, 24]  & $4$             \\
Twitter               & 2019      & 14.9        & 14         & [1, 169]     & [0, 24]  & $4$             \\
Yelp-Airport          & 319       & 30.5        & 7.5        & [9, 55]      & [0, 24]  & $4$             \\
Yelp-Miss.            & 319       & 55.2        & 15.9       & [3, 107]     & [0, 24]  & $4$             \\ \bottomrule
\end{tabular}
}
\end{table}

\clearpage

\section{Proofs}
\label{appendix:proofs}

In this section, we provide a proof of Proposition \ref{prop:balanced_couplings}, showing a necessary and sufficient condition for the existence of balanced couplings.

\balancedcouplings*
\begin{proof}
    Let $A_1, A_2 \subseteq \Gamma$ be Borel measurable \citep[Prop.~5.3]{daley2003introduction} subsets of the configuration space $\Gamma$, i.e. each of $A_1, A_2$ is a measurable collection of event sequences. Observe that for $i = 1, 2$, each $A_i$ can be written as a disjoint union
    \begin{equation} \label{eqn:ndim_proj}
        A_i^n = \bigcup_{n=0}^\infty \TT^n \cap A_i
    \end{equation}

    i.e. $A_i^n \subseteq A_i$ is the subset of $A_i$ containing only sequences with $n$ events. Note each $A_i^n$ is a Borel measurable subset of $\TT^n$.
    
    Now, suppose that $\mu(n) = \nu(n)$ have equal event count distributions. We define the coupling $\rho \in \P(\Gamma \times \Gamma)$ by
    \begin{equation} \label{eqn:indep_coupling}
        \rho(A_1 \times A_2) = \sum_{n=0}^\infty \mu(n) \mu^n(A_1^n) \nu^n(A_2^n).
    \end{equation}

    Here, in a slight abuse of notation, we use $\mu^n, \nu^n$ to denote the corresponding joint probability measures over $n$ events, i.e., both are Borel probability measures on $\TT^n$. Since the $n$-dimensional projection of $\Gamma$ in Equation \eqref{eqn:ndim_proj} is simply $\TT^n$, it is immediate that $\rho(A_1 \times \Gamma) = \mu(A_1)$ and $\rho(\Gamma \times A_2) = \nu(A_2)$, so that $\rho$ is indeed a coupling. Moreover, it is clear that the coupling is balanced.

    Conversely, suppose $\rho \in \Pi_b(\mu_0, \mu_1)$ is a balanced coupling. Let $N: \Gamma \to \Z_{\geq 0}$ be the event counting functional and let $\pi^1, \pi^2: \Gamma \times \Gamma \to \Gamma$ denote the canonical projections of $\Gamma \times \Gamma$ onto its components. That is, $\pi^1: (\gamma_0, \gamma_1) \mapsto \gamma_0$ and $\pi^2: (\gamma_0, \gamma_1) \mapsto \gamma_1$. Furthermore, let $(N, N): \Gamma \times \Gamma \to \Z_{\geq 0} \times \Z_{\geq 0}$ denote the product of the counting functional, i.e. $(N, N)(\gamma_0, \gamma_1) = (N(\gamma_0), N(\gamma_1))$. Note that the pushforward $N_{\#} \mu$ yields the event count distribution $\mu(n)$ of $\mu$ (and analogously for $\nu$). 
    
    Now, observe that composing the projections and counting functionals yields
    \begin{equation}
        \pi^1 \circ (N, N) = N \circ \pi^1 \qquad \pi^2 \circ (N, N) = N \circ \pi^2.
    \end{equation}
    As $\rho$ is a coupling, we have that $\mu = \pi^1_{\#}\rho$ and $\nu = \pi^2_{\#} \rho$. From these observations, it follows that
    \begin{align}
        N_{\#} \mu &= N_{\#} \left( \pi^1_{\#} \rho \right) \\
        &= (N \circ \pi^1)_{\#} \rho \\
        &= (\pi^1 \circ (N, N))_{\#}\rho \\
        &= \pi^1_{\#} \left( (N, N)_{\#} \rho\right) \\
        &= \pi^2_{\#} \left((N, N)_{\#} \rho\right) \\
        &= N_{\#}\nu
    \end{align}

    where the equality in the penultimate line follows from the fact that $\rho$ is balanced. Thus, we have shown that the existence of a balanced coupling implies that $N_{\#}\mu = N_{\#}\nu$, i.e. the event count distributions are equal.
\end{proof}

\section{EventFlow Architecture and Training Details}
\label{appendix:model_arch}
\looseness=-1
Here, we provide additional details regarding the parametrization and training of our \texttt{EventFlow} model. Our model is based on the transformer architecture \citep{vaswani2017attention, mei2022transformer}, due to its general ability to handle variable length inputs and outputs, high flexibility, and ability to incorporate long-range interactions. We emphasize that this is an effective architecture choice, but our method is not necessarily tied to this architecture. In all settings, our reference measure $\mu_0$ is specified with $q = \NN(0, I)$. All models (including the baselines) were trained on a cluster of six NVIDIA A6000 GPUs with 24Gb of VRAM. No hardware parallelization was used, i.e., each model was only trained on a single GPU.

\begin{figure}[p]
    \centering
    \includegraphics[width=4.8in, keepaspectratio]{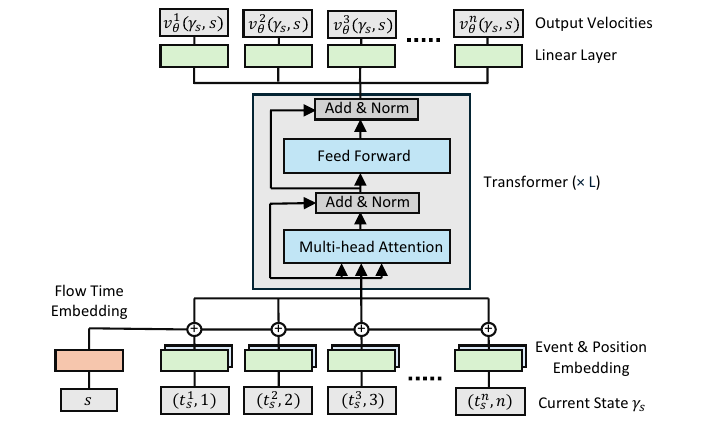}
    \caption{Overview of our model architecture for unconditional generation. The model takes as input the flow time $s$ and current sequence state $\gamma_s = \sum_{k=1}^n \delta[t_s^k]$. Each input is projected to a fixed-length vector via a learnable embedding. The resulting embeddings are  added and passed to the transformer model, which produces a sequence of output velocities $v_\theta(\gamma_s, s)$ with $N(\gamma_s)$ components.}
    \label{fig:uncond_arch}
\end{figure}

\looseness=-1
\paragraph{Model Parametrization}
Our unconditional model takes in a sequence $\gamma_s = (t^1_s, \dots, t^n_s)$ and flow time $s$. We first embed the sequence times $t^k$, the flow-time $s$, and the sequence position indices. These position indices are handled by sinusoidal embeddings followed by an additional linear layer. There are three linear layers in total: one for the flow time, one shared across the sequence times, and one for the position indices. These embeddings are added together to create a representation for each element of the sequence, and we apply a standard transformer to this sequence to produce a sequence of vectors of length $N(\gamma_s)$. Finally, each of these vectors is projected to one dimension via a final linear layer with shared weights to produce the vector field $v_\theta(\gamma_s, s)$. See Figure \ref{fig:uncond_arch}.

For the conditional model, we use a standard transformer encoder-decoder architecture. We first embed the history sequence times $\HH$ and the sequence position indices in a manner analogous to the above. The model was provided the start of the prediction window $T_0$ by concatenating it as the final event in $\HH$. This yielded better results than encoding the start of the prediction window separately. We feed these embeddings through the transformer encoder produce an intermediate representation $e_{\mathcal{H}}$. 

\begin{figure}[!t]
    \centering
    \includegraphics[width=5.5in, keepaspectratio]{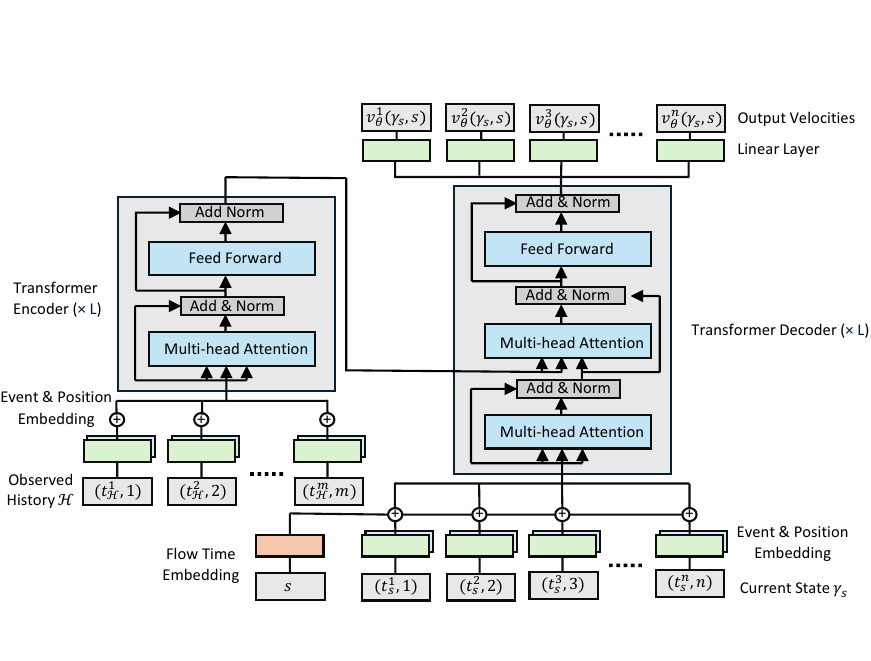}
    \caption{Overview of our model architecture for conditional generation. The encoder (left) takes as input the observed history $\HH$, which is embedded in a fashion analogous to our unconditional model. The decoder (right) takes as input the flow time $s$ and current state $\gamma_s = \sum_{k=1} \delta[t_s^k]$. These are embedded and passed through the decoder, which applies cross attention to produce the conditional velocities $v_\theta(\gamma_s, s, e_\HH)$.}
    \label{fig:cond_arch}
\end{figure}

For the decoder, we provide the model with the current state $\gamma_s$ (corresponding to the generated event times at flow-time $s$), the flow-time $s$, and the corresponding positional indices. These are embedded as previously described, before being passed into the transformer decoder. The history encoding $e_{\HH}$ is provided to the decoder via cross-attention in the intermediate layer. This produces a sequence of $N(\gamma_s)$ vectors, which we again pass through a final linear layer to produce the final conditional vector field $v_\theta(\gamma_s, s, e_{\HH})$. See Figure \ref{fig:cond_arch}.

\begin{figure}[!t]
    \centering
    \includegraphics[width=4.8in, keepaspectratio]{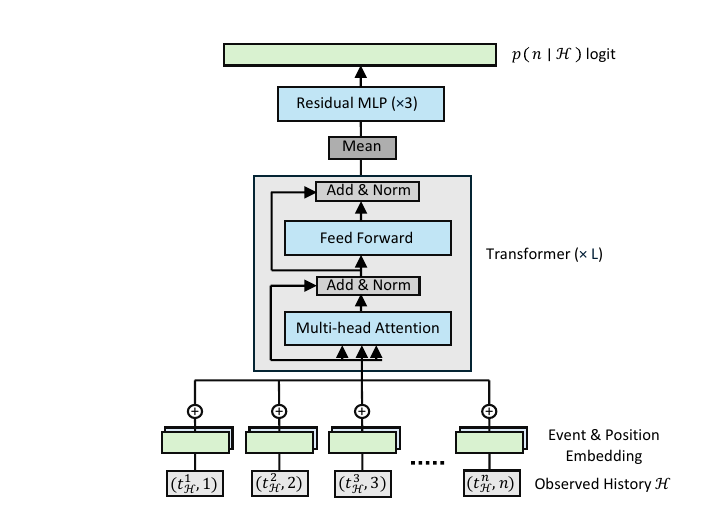}
    \caption{Overview of our architecture modeling the event count distribution $p_\phi(n \mid \HH)$. The model takes as input an observed history $\HH$. As in our other architectures, the events are embedded and passed through a transformer. Here, the final sequence embedding output by the transformer is averaged and passed through an additional residual MLP with three layers to produce the logit corresponding to $p_\phi(n \mid \HH)$.}
    \label{fig:cond_classifier}
\end{figure}

Our architecture for predicting the number of future events given a history, i.e. $p_\phi(n \mid \HH)$, is again based on the transformer decoder, sharing the same overall architecture as our unconditional model. The key difference is that we instead take a mean of the final sequence embeddings before passing this through a small MLP to produce the final logit. See Figure \ref{fig:cond_classifier}. We note that this requires specifying a maximum value $N_{\max}$, which we set as the maximum sequence length seen in the training data. In early experiments, we also parametrized the model not through a logit but rather as a mixture of Poissons or a mixture of Negative Binomials. While this no longer has the limitation of assuming a value for $N_{\max}$, we found that this approach yielded worse results. 

\paragraph{Training and Tuning}

We normalize all sequences to the range $[-1, 1]$, using the overall min/max event time seen in the training data. All sequences are generated on this normalized scale, prior to re-scaling the sequence back to the original data range before evaluation. Our models are trained with the Adam \citep{kingma2017adammethodstochasticoptimization} optimizer with $\beta_1 = 0.9$ and $\beta_2 = 0.999$ for $30,000$ steps with a cosine scheduler \citep{loshchilov2017sgdrstochasticgradientdescent}, which cycled every $10,000$ steps. Final hyperparameters were selected by best performance on the validation dataset achieved at any point during the training, where models were evaluated 10 times throughout their training.

To tune our model, we performed a grid search over learning rates in $\{ 5 \times 10^{-3}, 10^{-3}, 5 \times 10^{-4} \}$ and dropout probabilities in $\{ 0, 0.1, 0.2 \}$. Overall, we found that learning rates of $10^{-2}$ or larger often caused the model to diverge. We use $6$ transformer layers, $8$ attention heads, and an embedding dimension of $512$ across all settings, except for the Reddit-C and Reddit-S datasets where we use $4$ heads and an embedding dimension of $128$ due to the increased memory cost of these datasets.

To train the event count model $p_\phi(n \mid \HH)$, we seek to minimize the regularized cross-entropy loss
\begin{equation}
    \mathcal{L}(\phi) = \E_{n, \HH} \left[- \log p_\phi(n \mid \HH) + \frac{\alpha}{N_{\max}} \sum_{k=1}^{N_{\max}} p_\phi(k \mid \HH) (n - k)^2 \right]
\end{equation}
where $\HH$ is a given history and $n$ is the number of events in $\TT$ following this history. Note that this is a cross-entropy loss where the regularization term can be viewed as a squared optimal transport distance between the distribution $p_\phi(n \mid \HH)$ and a delta distribution $\delta[n]$ at the true $n$. This regularization term encourages the values of $p_\phi(n \mid \HH)$ to be smooth in $n$, as it penalizes predictions which are far fronm the true $n$ more heavily than those which are nearby. We searched for values of $\alpha$, the weight associated with the OT-loss, over the set $\{0, 1/N_{\max}, 10/N_{\max}, 100/N_{\max}, 1000/N_{\max} \}$.

\vspace{-5cm}

\clearpage
\begin{table}[!t]
\caption{The best hyperparameter settings found for the vector field $v_\theta$ in our \texttt{EventFlow} method on the unconditional generation task.}
\vspace{1em}
\centering
\small{
\begin{tabular}{@{}lcccccc@{}}
\toprule
                      & Learning Rate & Emb. Dim. & MLP Dim & Heads & Transformer Layers & Dropout\\ \midrule
Hawkes1               & $10^{-3}$     & $512$ & $2048$ & $8$ & $6$ & 0.1                     \\
Hawkes2               & $10^{-3}$     & $512$ & $2048$ &$8$ & $6$  & 0.1                             \\
Nonstationary Poisson & $10^{-3}$     & $512$ & $2048$ &$8$ & $6$ & 0.1                               \\
Nonstationary Renewal & $10^{-3}$     & $512$ & $2048$ &$8$ & $6$ & 0.1                                 \\
Stationary Renewal    & $10^{-3}$     & $512$ & $2048$ &$8$ & $6$ & 0.1                                 \\
Self-Correcting       & $10^{-3}$     & $512$ & $2048$ &$8$ & $6$ & 0.1                                \\ \midrule
PUBG                  & $5\times10^{-4}$     & $512$ & $2048$ &$8$ & $6$ & 0.1                                \\
Reddit-C              & $10^{-3}$     & $128$ & $256$ &$4$ & $6$ & 0.1                              \\
Reddit-S              & $5\times10^{-3}$     & $128$ &$256$ & $4$ & $6$ & 0.1                              \\
Taxi                  & $5\times10^{-4}$     & $512$ & $2048$ &$8$ & $6$ & 0.1                             \\
Twitter               & $10^{-3}$     & $512$ & $2048$ &$8$ & $6$ & 0.1                            \\
Yelp-Airport          & $5\times10^{-4}$     & $512$ & $2048$ &$8$ & $6$ & 0.1                               \\
Yelp-Miss.            & $10^{-3}$     & $512$ & $2048$ &$8$ & $6$ & 0.1                           \\ \bottomrule
\end{tabular}
}
\end{table}

\begin{table}[!hbt]
\caption{The best hyperparameter settings found for the vector field $v_\theta$ in our \texttt{EventFlow} method on the forecasting task.}
\vspace{1em}
\centering
\small{
\begin{tabular}{@{}lcccccc@{}}
\toprule
                      & Learning Rate & Emb. Dim. & MLP Dim. & Heads & Transformer Layers & Dropout         \\ \midrule
PUBG                  & $5\times10^{-3}$     & $512$ & $2048$ &$8$ & $6$ & $0.1$                              \\
Reddit-C              & $10^{-3}$     & $128$ & $256$ &$4$ & $6$ & $0.2$                               \\
Reddit-S              & $10^{-3}$     & $128$ & $256$ & $4$ & $6$ & $0.2$                            \\
Taxi                  & $5\times10^{-4}$     & $512$ & $2048$ &$8$ & $6$ & $0.1$                            \\
Twitter               & $5\times10^{-4}$     & $512$ &$2048$ & $8$ & $6$ & $0.1$                           \\
Yelp-Airport          & $10^{-3}$     & $512$ & $2048$ &$8$ & $6$ & $0.2$                              \\
Yelp-Miss.            & $5\times10^{-4}$     & $512$ &$2048$ & $8$ & $6$ & $0.1$                          \\ \bottomrule
\end{tabular}
}
\end{table}

\begin{table}[!hbt]
\caption{The best hyperparameter settings found for the event count predictor $p_\phi(n \mid \HH)$ in our \texttt{EventFlow} method on the forecasting task.}
\vspace{1em}
\centering
\small{
\begin{tabular}{@{}lccccccc@{}}
\toprule
                      & Learning Rate & $\alpha / N_{max}$ & Emb. Dim.  & MLP Dim. & Heads & Transformer Layers & Dropout        \\ \midrule
PUBG                  & $10^{-3}$   & $1.0$ & $512$ & $2048$ & $8$ & $6$ & $0.2$                              \\
Reddit-C              & $10^{-3}$   & $1000.0$ & $128$ & $256$ &$4$ & $6$ & $0.2$                             \\
Reddit-S              & $10^{-3}$   & $1000.0$ & $128$ & $256$ &$4$ & $6$ & $0.0$                            \\
Taxi                  & $10^{-3}$   & $1000.0$ & $512$ & $2048$ &$8$ & $6$ & $0.2$                            \\
Twitter               & $10^{-3}$   & $1000.0$& $512$ & $2048$ &$8$ & $6$ & $0.2$                           \\
Yelp-Airport          & $10^{-3}$   & $1000.0$& $512$ & $2048$ &$8$ & $6$ & $0.2$                              \\
Yelp-Miss.            & $10^{-3}$   & $100.0$& $512$ & $2048$ &$8$ & $6$ & $0.2$                          \\ \bottomrule
\end{tabular}
}
\end{table}

\clearpage

\section{Additional Details on Baselines}
\label{appendix:baselines}

In this section, we provide additional details regarding our baseline methods. All methods are trained at a batch size of $64$ for $1,000$ epochs, using early stopping on the validation set loss. In early experiments, we also evaluated AttNHP \citep{zuo2020transformer}, a variant of the NHP which uses an attention-based encoder, but found it to be prohibitively expensive in terms of memory cost (requiring more than 24 GB of VRAM) and, as a result, do not include it in our results. 

\paragraph{IFTPP} Our first baseline is the intensity-free TPP model of \citet{shchurintensity}. This model uses an RNN encoder and a mixture of log-normal distributions to parametrize the decoder. We directly use the implementation provided by the authors.\footnote{URL: \href{https://github.com/shchur/ifl-tpp}{\texttt{https://github.com/shchur/ifl-tpp}}}. We train for $1,000$ epochs with early stopping based on the validation set loss. To tune this baseline, we performed a grid search over learning rates in $\{ 10^{-4}, 10^{-3}, 10^{-2} \}$, weight decays in $\{ 0, 10^{-6}, 10^{-5}, 10^{-4} \}$, history embedding dimensions $\{ 32, 64, 128 \}$, and mixture component counts $\{8, 16, 32, 64 \}$. Our best hyperparameters can be found in Table \ref{tab:iftpp_hps_uncond} and Table \ref{tab:iftpp_hps_cond}. 

\begin{table}[!hbt]
\caption{The best hyperparameter settings found for IFTPP on the unconditional generation task.}
\label{tab:iftpp_hps_uncond}
\vspace{1em}
\centering
\small{
\begin{tabular}{@{}lcccc@{}}
\toprule
                      & Learning Rate & Weight Decay & Embedding Dimension & Mixture Components  \\ \midrule
Hawkes1               & $10^{-3}$     & $10^{-4}$    & $32$                & $8$                 \\
Hawkes2               & $10^{-2}$     & $0$          & $32$                & $8$                 \\
Nonstationary Poisson & $10^{-3}$     & $10^{-6}$    & $128$               & $8$                 \\
Nonstationary Renewal & $10^{-2}$     & $10^{-6}$    & $64$                & $16$                \\
Stationary Renewal    & $10^{-3}$     & $10^{-4}$    & $32$                & $8$                 \\
Self-Correcting       & $10^{-3}$     & $10^{-6}$    & $32$                & $64$                \\ \midrule
PUBG                  & $10^{-2}$     & $0$          & $128$               & $32$                \\
Reddit-C              & $10^{-3}$     & $10^{-4}$    & $64$                & $16$                \\
Reddit-S              & $10^{-2}$     & $10^{-4}$    & $64$                & $16$                \\
Taxi                  & $10^{-2}$     & $10^{-5}$    & $128$               & $64$                \\
Twitter               & $10^{-3}$     & $10^{-4}$    & $64$                & $6$                 \\
Yelp-Airport          & $10^{-2}$     & $10^{-6}$    & $64$                & $64$                \\
Yelp-Miss.            & $10^{-3}$     & $10^{-4}$    & $32$                & $8$                 \\ \bottomrule
\end{tabular}
}
\end{table}

\begin{table}[!hbt]
\caption{The best hyperparameter settings found for IFTPP on the forecasting task.}
\label{tab:iftpp_hps_cond}
\vspace{1em}
\centering
\small{
\begin{tabular}{@{}lcccc@{}}
\toprule
                      & Learning Rate & Weight Decay & Embedding Dimension & Mixture Components  \\ \midrule
PUBG                  & $10^{-4}$     & $10^{-6}$    & $32$                & $32$                \\
Reddit-C              & $10^{-2}$     & $0$          & $64$                & $8$                 \\
Reddit-S              & $10^{-2}$     & $0$          & $64$                & $16$                \\
Taxi                  & $10^{-3}$     & $10^{-6}$    & $128$               & $8$                 \\
Twitter               & $10^{-2}$     & $10^{-5}$    & $32$                & $8$                 \\
Yelp-Airport          & $10^{-2}$     & $10^{-6}$    & $128$               & $32$                \\
Yelp-Miss.            & $10^{-2}$     & $10^{-6}$    & $32$                & $8$                 \\ \bottomrule
\end{tabular}
}
\end{table}

\clearpage
\FloatBarrier

\paragraph{NHP} We additionally compare against the Neural Hawkes Process of \citet{mei2017neural}. This model uses an LSTM encoder and a parametric form, whose weights are modeled by a neural network, to model the conditional intensity function. In practice, we use the implementation proved by the EasyTPP benchmark \citep{xue2023easytpp}, as this version implements the necessary thinning algorithm for sampling.\footnote{URL: \href{https://github.com/ant-research/EasyTemporalPointProcess}{\texttt{https://github.com/ant-research/EasyTemporalPointProcess}}} We perform a grid search over learning rates in $\{ 10^{-4}, 10^{-3}, 10^{-2} \}$ and embedding dimensions in $\{32, 64, 128 \}$. These hyperparameters are chosen as the EasyTPP implementation allows these to be configured easily. Our best hyperparameters are reported in Table \ref{tab:nhp_hps_uncond} and Table \ref{tab:nhp_hps_cond}.

\begin{table}[!hbt]
\caption{The best hyperparameter settings found for NHP on the unconditional generation task.}
\label{tab:nhp_hps_uncond}
\vspace{1em}
\centering
\small{
\begin{tabular}{@{}lll@{}}
\toprule
                      & Learning Rate & Embedding Dimension       \\ \midrule
Hawkes1               & $10^{-3}$     & $64$                      \\
Hawkes2               & $10^{-3}$     & $64$                      \\
Nonstationary Poisson & $10^{-3}$     & $64$                      \\
Nonstationary Renewal & $10^{-4}$     & $64$                      \\
Stationary Renewal    & $10^{-3}$     & $64$                      \\
Self-Correcting       & $10^{-3}$     & $64$                      \\ \midrule
PUBG                  & $10^{-4}$     & $64$                      \\
Reddit-C              & $10^{-2}$     & $64$                      \\
Reddit-S              & $10^{-2}$     & $64$                      \\
Taxi                  & $10^{-2}$     & $64$                      \\
Twitter               & $10^{-4}$     & $64$                      \\
Yelp-Airport          & $10^{-3}$     & $128$                     \\
Yelp-Miss.            & $10^{-2}$     & $64$                      \\ \bottomrule
\end{tabular}
}
\end{table}

\begin{table}[!hbt]
\caption{The best hyperparameter settings found for NHP on the forecasting task.}
\label{tab:nhp_hps_cond}
\centering
\small{
\begin{tabular}{@{}lll@{}}
\toprule
                      & Learning Rate & Embedding Dimension       \\ \midrule
PUBG                  & $10^{-3}$     & $128$                     \\
Reddit-C              & $10^{-2}$     & $64$                      \\
Reddit-S              & $10^{-2}$     & $64$                      \\
Taxi                  & $10^{-2}$     & $128$                     \\
Twitter               & $10^{-2}$     & $128$                     \\
Yelp-Airport          & $10^{-3}$     & $64$                      \\
Yelp-Miss.            & $10^{-2}$     & $64$                      \\ \bottomrule
\end{tabular}
}
\end{table}

\FloatBarrier

\paragraph{Diffusion} Our diffusion baseline is based on the implementation of \citet{lin2022exploring}, and our decoder model architecture is taken directly from the code of \citet{lin2022exploring}.\footnote{URL: \href{https://github.com/EDAPINENUT/GNTPP}{\texttt{https://github.com/EDAPINENUT/GNTPP}}} At a high level, this model is a discrete-time diffusion model \citep{ho2020denoising} trained to generate a single inter-arrival time given a history embedding. Note that as the likelihood is not available in diffusion models, the CDF in the likelihood in Equation \eqref{eqn:likelihood} is not tractable. Instead, the model is trained by maximizing an ELBO of only the subsequent inter-arrival time.

In preliminary experiments, we found that the codebase provided by \citet{lin2022exploring} often produced \texttt{NaN} values during sampling, prompting us to make several changes. First, we use the RNN encoder from \citet{shchurintensity}, i.e. the same encoder as the IFTPP baseline, to reduce the memory requirements of the model. Second, we do not log-scale the inter-arrival times as suggested by \citet{lin2022exploring}, as we found that this often led to overflow and underflow issues at sampling time. Third, we do not normalize the data via standardization (i.e., subtracting off the mean inter-arrival time and dividing by the standard deviation), but rather, we scale the inter-arrival times so that they are in the bounded range $[-1, 1]$. This is aligned with standard diffusion implementations \citep{ho2020denoising}, and allows us to perform clipping at sampling time to avoid the accumulation of errors. With these changes, our diffusion baseline is competitive, and able to obtain stronger results than previous work has reported \citep{ludke2023add}. 

We use $1000$ diffusion steps and the cosine beta schedule \citep{nichol2021improved}, and we train the model on the simplified $\epsilon$-prediction loss of \citet{ho2020denoising}. We train for $1,000$ epochs with early stopping based on the validation set loss. To tune this baseline, we performed a grid search over learning rates in $\{ 10^{-4}, 10^{-3}, 10^{-2} \}$, weight decays in $\{ 0, 10^{-6}, 10^{-5}, 10^{-4} \}$, history embedding dimensions $\{ 32, 64, 128 \}$, and layer numbers $\{2, 4, 6\}$. Our best hyperparameters can be found in Table \ref{tab:diff_hps_uncond} and Table \ref{tab:diff_hps_cond}.

\begin{table}[!hbt]
\caption{The best hyperparameter settings found for diffusion on the unconditional generation task.}
\label{tab:diff_hps_uncond}
\centering
\vspace{1em}
\small{
\begin{tabular}{@{}lllll@{}}
\toprule
                      & Learning Rate & Weight Decay & Embedding Dimension & Layers        \\ \midrule
Hawkes1               & $10^{-3}$     & $10^{-6}$    & $64$                & $2$           \\
Hawkes2               & $10^{-2}$     & $10^{-5}$    & $64$                & $4$           \\
Nonstationary Poisson & $10^{-3}$     & $10^{-5}$    & $128$               & $2$           \\
Nonstationary Renewal & $10^{-3}$     & $10^{-4}$    & $64$                & $2$           \\
Stationary Renewal    & $10^{-2}$     & $0$          & $32$                & $6$           \\
Self-Correcting       & $10^{-3}$     & $0$          & $32$                & $6$           \\ \midrule
PUBG                  & $10^{-3}$     & $0$          & $64$                & $2$           \\
Reddit-C              & $10^{-3}$     & $10^{-6}$    & $128$               & $4$           \\
Reddit-S              & $10^{-3}$     & $0$          & $64$                & $4$           \\
Taxi                  & $10^{-2}$     & $0$          & $128$               & $4$           \\
Twitter               & $10^{-3}$     & $10^{-4}$    & $64$                & $6$           \\
Yelp-Airport          & $10^{-2}$     & $0$          & $32$                & $2$           \\
Yelp-Miss.            & $10^{-2}$     & $10^{-5}$    & $128$               & $2$           \\ \bottomrule
\end{tabular}
}
\end{table}

\begin{table}[!hbt]
\caption{The best hyperparameter settings found for diffusion on the forecasting task.}
\label{tab:diff_hps_cond}
\centering
\vspace{1em}
\small{
\begin{tabular}{@{}lllll@{}}
\toprule
                      & Learning Rate & Weight Decay & Embedding Dimension & Layers        \\ \midrule
PUBG                  & $10^{-4}$     & $10^{-5}$    & $32$                & $6$      \\
Reddit-C              & $10^{-2}$     & $10^{-6}$    & $64$                & $6$           \\
Reddit-S              & $10^{-3}$     & $0$          & $64$                & $4$           \\
Taxi                  & $10^{-3}$     & $10^{-6}$    & $32$                & $2$           \\
Twitter               & $10^{-4}$     & $10^{-5}$    & $64$                & $6$           \\
Yelp-Airport          & $10^{-4}$     & $10^{-5}$    & $64$                & $6$           \\
Yelp-Miss.            & $10^{-3}$     & $10^{-5}$    & $32$                & $4$           \\ \bottomrule
\end{tabular}
}
\end{table}

\FloatBarrier
\clearpage

\paragraph{Add-and-Thin} We compare to the \texttt{Add-and-Thin} model of \citet{ludke2023add} as a recently proposed non-autoregressive baseline. We directly run the code provided by the authors without additional modifications.\footnote{URL: \href{https://github.com/davecasp/add-thin}{\texttt{https://github.com/davecasp/add-thin}}} We do, however, perform a slightly larger hyperparameter sweep than \citet{ludke2023add}, in order to ensure a fair comparison between the methods considered. We train for $1,000$ epochs with early stopping on the validation loss. Tuning is performed via a grid search over learning rates in $\{ 10^{-4}, 10^{-3}, 10^{-2} \}$ and number of mixture components in $\{ 8, 16, 32, 64 \}$. We choose to tune only these hyperparameters in order to follow the implementation provided by the authors. Our best hyperparameters can be found in Table \ref{tab:addthin_hps_uncond} and Table \ref{tab:addthin_hps_cond}.

\begin{table}[!hbt]
\caption{The best hyperparameter settings found for Add-and-Thin on the unconditional generation task.}
\label{tab:addthin_hps_uncond}
\centering
\vspace{1em}
\small{
\begin{tabular}{@{}lcc@{}}
\toprule
                      & Learning Rate & Mixture Components        \\ \midrule
Hawkes1               & $10^{-3}$     & $32$                      \\
Hawkes2               & $10^{-2}$     & $32$                      \\
Nonstationary Poisson & $10^{-2}$     & $16$                      \\
Nonstationary Renewal & $10^{-2}$     & $8$                       \\
Stationary Renewal    & $10^{-2}$     & $8$                       \\
Self-Correcting       & $10^{-4}$     & $8$                       \\ \midrule
PUBG                  & $10^{-3}$     & $8$                       \\
Reddit-C              & $10^{-2}$     & $32$                      \\
Reddit-S              & $10^{-2}$     & $16$                      \\
Taxi                  & $10^{-2}$     & $8$                       \\
Twitter               & $10^{-4}$     & $32$                      \\
Yelp-Airport          & $10^{-4}$     & $8$                       \\
Yelp-Miss.            & $10^{-2}$     & $64$                      \\ \bottomrule
\end{tabular}
}
\end{table}

\begin{table}[!htb]
\caption{The best hyperparameter settings found for Add-and-Thin on the forecasting task.}
\label{tab:addthin_hps_cond}
\centering
\vspace{1em}
\small{
\begin{tabular}{@{}lcc@{}}
\toprule
                      & Learning Rate & Mixture Components        \\ \midrule
PUBG                  & $10^{-2}$     & $64$                      \\
Reddit-C              & $10^{-2}$     & $16$                      \\
Reddit-S              & $10^{-2}$     & $64$                      \\
Taxi                  & $10^{-2}$     & $8$                       \\
Twitter               & $10^{-3}$     & $8$                       \\
Yelp-Airport          & $10^{-2}$     & $32$                      \\
Yelp-Miss.            & $10^{-3}$     & $16$                      \\ \bottomrule
\end{tabular}
}
\end{table}

\clearpage
\FloatBarrier

\section{Additional Experiments}
\label{appendix:additional_experiments}

This section contains additional empirical evaluations, supplementing Section \ref{sec:experiments}.

\subsection{MARE}

 In Table \ref{tab:mape_with_sddev}, we evaluate the performance of the various models only in terms of the predicted number of events in the forecast. To do so, we measure the mean absolute relative error (MARE) given by
\begin{equation}
    \mathrm{MARE} = \mathbb{E}_{n, \HH, \hat{n}\sim p_\phi(n \mid \HH)}\left| \frac{\hat{n} - n}{n} \right|
\end{equation}

where $n$ represents the true number of points in a sequence following a history $\HH$ and  $\hat{n} \sim p_\phi(n \mid \HH)$ represents the predicted number of points. The expectation is estimated empirically on the testing set. As our method directly predicts the number of events $n$ by sampling from the learned distribution $p_\phi(n \mid \HH)$, this serves as a direct evaluation of this component of our model. We note that the baselines only model $n$ implicitly. For example, the autoregressive models sample new events until an event is generated outside the support $\TT$. As an ablation, we also report the MARE values for our method without regularization, i.e., setting $\alpha = 0$. We see that the regularization significantly improves the event count predictions.

\begin{table}[!ht]
\caption{MARE values evaluating the predicted number of events when forecasting. Mean values $\pm$ one standard deviation are reported over five random seeds. The lowest MARE on each dataset is indicated and bold, and the second lowest is indicated by an underline.}
\label{tab:mape_with_sddev}
\centering
\vspace{1em}
\small{
\begin{tabular}{@{}lrrrrrrr@{}}
\toprule
                      & PUBG & Reddit-C & Reddit-S & Taxi & Twitter & Yelp-A & Yelp-M \\ \midrule
                IFTPP & $1.05 {\scriptstyle \pm 0.14}$ & $1.69 {\scriptstyle \pm 0.39}$ & $0.79 {\scriptstyle \pm 0.20}$ & $0.60 {\scriptstyle \pm 0.11}$ & $0.88 {\scriptstyle \pm 0.08}$ & $0.76 {\scriptstyle \pm 0.07}$ & $0.76 {\scriptstyle \pm 0.05}$ \\ 
                NHP   & $1.02 {\scriptstyle \pm 0.08}$ & $\underline{0.95} {\scriptstyle \pm 0.01}$ & $1.00 {\scriptstyle \pm 0.0004}$ & $0.67 {\scriptstyle \pm 0.11}$ & $2.48 {\scriptstyle \pm 0.40}$ & $0.80 {\scriptstyle \pm 0.22}$ & $1.07 {\scriptstyle \pm 0.34}$ \\
            Diffusion & $1.95 {\scriptstyle \pm 0.48}$  & $1.28 {\scriptstyle \pm 0.09}$ & $1.12 {\scriptstyle \pm 0.56}$ & $0.49 {\scriptstyle \pm 0.07}$ & $0.66 {\scriptstyle \pm 0.04}$ & $0.65  {\scriptstyle \pm 0.07}$ & $0.72 {\scriptstyle \pm 0.07} $\\
         Add-and-Thin & $\underline{0.43} {\scriptstyle \pm 0.01}$ & $0.99 {\scriptstyle \pm 0.10} $ & $\underline{0.38} {\scriptstyle \pm 0.01} $ & $\underline{0.33} {\scriptstyle \pm 0.02}$ & $\underline{0.60} {\scriptstyle \pm 0.02}$ & $\mathbf{0.42} {\scriptstyle \pm 0.01}$ &  $\mathbf{0.46} {\scriptstyle \pm 0.03}$\\
                 EventFlow (ours) & $\mathbf{0.40} {\scriptstyle \pm 0.01}$ &  $\mathbf{0.70} {\scriptstyle \pm 0.07}$ & $\mathbf{0.16} {\scriptstyle \pm 0.01} $& $\mathbf{0.28} {\scriptstyle \pm 0.01}$ & $\mathbf{0.46} {\scriptstyle \pm 0.03}$ & $\underline{0.56} {\scriptstyle \pm 0.06}$ & $ \underline{0.50} {\scriptstyle \pm 0.02}$ \\
                 \midrule
                 EventFlow ($\alpha = 0$) & $0.69 {\scriptstyle \pm 0.17}$ & $2.01 {\scriptstyle \pm 0.40}$ & $0.26 {\scriptstyle \pm 0.01}$ & $0.47 {\scriptstyle \pm 0.03}$ & $1.23 {\scriptstyle \pm 0.07}$ & $0.66 {\scriptstyle \pm 0.03}$ & $0.80 {\scriptstyle \pm 0.05}$ \\
                 \bottomrule
                 
\end{tabular}
}
\end{table}

\FloatBarrier

\subsection{MMDs}
In Tables \ref{tab:mmd_real} and \ref{tab:mmd_synthetic}, we report the MMD values appearing in the unconditional experiment (i.e., Table \ref{tab:mmd_all} in the main paper) with standard deviations. These are omitted from the main paper for the sake of space. 

\begin{table}[!h] 
\caption{MMDs (1e-2) between the test set and $1,000$ generated sequences on our real-world datasets. Lower is better. We report the mean $\pm$ one standard deviation over five random seeds. The lowest MMD distance on each dataset is indicated in bold, and the second lowest is indicated by an underline.}
\label{tab:mmd_real}
\centering
\small{
\begin{tabular}{@{}lrrrrrrr@{}}
\toprule
                      & PUBG & Reddit-C & Reddit-S & Taxi & Twitter & Yelp-A & Yelp-M \\ \midrule
                Data & $1.3$ & $0.6$ & $0.4$ & $3.1$ & $2.6$ & $3.6$ & $3.1$ \\ \midrule
                IFTPP & $5.7 {\scriptstyle \pm 1.8}$ & $1.3 {\scriptstyle \pm 1.2}$ & $\underline{1.9} {\scriptstyle \pm 0.6}$ & $5.8 {\scriptstyle \pm 0.9}$ & $\mathbf{2.9} {\scriptstyle \pm 0.6}$ & $8.2 {\scriptstyle \pm 4.7}$ & $\underline{5.1} {\scriptstyle \pm 0.7}$ \\ 
                NHP   & $7.2 {\scriptstyle \pm 0.2}$ & $2.2 {\scriptstyle \pm 1.6}$ & $22.5 {\scriptstyle \pm 0.3}$ & $\underline{5.0} {\scriptstyle \pm 0.1}$ & $7.3 {\scriptstyle \pm 0.7}$ & $6.7 {\scriptstyle \pm 1.5}$ & $6.1 {\scriptstyle \pm 2.3}$ \\
            Diffusion & $14.3 {\scriptstyle \pm 6.5}$ & $3.9 {\scriptstyle \pm 1.2}$ & $6.2 {\scriptstyle \pm 3.3}$ & $11.7 {\scriptstyle \pm 1.8}$ & $12.5 {\scriptstyle \pm 1.9}$ & $10.9 {\scriptstyle \pm 3.8}$ & $10.5 {\scriptstyle \pm 5.2}$ \\
         Add-and-Thin & $\underline{2.8} {\scriptstyle \pm 0.5}$ & $\underline{1.2} {\scriptstyle \pm 0.27}$ & $2.7 {\scriptstyle \pm 0.3}$ & $
         5.2 {\scriptstyle \pm 0.6}$ & $\underline{4.8} {\scriptstyle \pm 0.4}$ & $\mathbf{4.5} {\scriptstyle \pm 0.2}$ & $\mathbf{3.0} {\scriptstyle \pm 0.5}$ \\
                 EventFlow (ours) & $\mathbf{1.5} {\scriptstyle \pm 0.2}$ & $\mathbf{0.7} {\scriptstyle \pm 0.1} $& $\mathbf{0.7} {\scriptstyle \pm 0.1} $& $\mathbf{3.5} {\scriptstyle \pm 0.1}$ & $4.9 {\scriptstyle \pm 0.7}$ & $\underline{6.6} {\scriptstyle \pm 1.2}$ & $\mathbf{3.0} {\scriptstyle \pm 0.5}$ \\ \bottomrule

\end{tabular}
}
\end{table}

\begin{table}[!h] 
\caption{MMDs (1e-2) between the test set and $1,000$ generated sequences on our synthetic datasets. Lower is better. We report the mean $\pm$ one standard deviation over five random seeds. The lowest MMD distance on each dataset is indicated in bold, and the second lowest is indicated by an underline.}
\label{tab:mmd_synthetic}
\vspace{1em}
\centering
\small{
\begin{tabular}{@{}lrrrrrr@{}}
\toprule
                      & Hawkes1 & Hawkes2 & NSP   & NSR   & SC    & SR    \\ \midrule
                Data  & $1.3$   & $1.3$   & $1.8$ & $3.0$ & $5.7$ & $1.1$ \\ \midrule
                IFTPP & $\mathbf{1.5} {\scriptstyle \pm 0.4}$ & $\mathbf{1.4} {\scriptstyle \pm 0.5}$ & $\mathbf{2.3} {\scriptstyle \pm 0.7}$ & $\underline{6.2} {\scriptstyle \pm 2.1}$ & $\mathbf{5.8} {\scriptstyle \pm 0.5}$ & $\mathbf{1.3} {\scriptstyle \pm 0.3}$ \\
                NHP   & $\underline{1.9} {\scriptstyle \pm 0.3}$ & $5.2 {\scriptstyle \pm 1.6}$ & $3.6 {\scriptstyle \pm 1.3}$ & $12.6 {\scriptstyle \pm 1.8}$ & $25.4 {\scriptstyle \pm 11.5}$ & $5.0 {\scriptstyle \pm 0.7}$  \\
            Diffusion & $4.8 {\scriptstyle \pm 2.7}$ & $5.5 {\scriptstyle \pm 3.3}$ & $10.8 {\scriptstyle \pm 7.5}$ & $15.0 {\scriptstyle \pm 3.6}$ & $9.1 {\scriptstyle \pm 1.8}$ & $5.1 {\scriptstyle \pm 2.8}$ \\
         Add-and-Thin & $\underline{1.9} {\scriptstyle \pm 0.5}$ & $2.5 {\scriptstyle \pm 0.3}$ & $\underline{2.6} {\scriptstyle \pm 0.5}$ & $7.4 {\scriptstyle \pm 1.2}$ & $22.5 {\scriptstyle \pm 0.5}$ & $2.2 {\scriptstyle \pm 0.8}$ \\
                 EventFlow (ours) & $\underline{1.9} {\scriptstyle \pm 0.2}$ & $\underline{2.2} {\scriptstyle \pm 0.1}$ & $3.8 {\scriptstyle \pm 1.2}$ & $\mathbf{4.2} {\scriptstyle \pm 0.5}$ & $\underline{8.3} {\scriptstyle \pm 0.4}$ & $\underline{1.7} {\scriptstyle \pm 0.3}$ \\ \bottomrule

\end{tabular}
}
\end{table}

\FloatBarrier

\subsection{MSEs}
\looseness=-1
In Table \ref{tab:mse}, we evaluate the performance of our model when forecasting only a single subsequent event. That is, given a history $\HH$, we evaluate the MSE between the first true event time following this history and the first event time generated by each model conditioned on $\HH$. The results are reported in Table \ref{tab:mse}. Generally, all of the methods show similar results on this metric, despite there being clear differences between methods on the multi-step task. We believe this serves to further highlight the necessity of moving beyond single-step prediction tasks.

\begin{table}[!h]
\caption{MSE values evaluating one-step-ahead forecasting performance. Mean values $\pm$ one standard deviation are reported over five random seeds. The lowest MSE on each dataset is indicated and bold, and the second lowest is indicated by an underline.}
\label{tab:mse}
\centering
\vspace{1em}
\small{
\begin{tabular}{@{}lrrrrrrr@{}}
\toprule
                      & PUBG & Reddit-C & Reddit-S & Taxi & Twitter & Yelp-A & Yelp-M \\ \midrule
                IFTPP & $0.85 {\scriptstyle \pm 0.05}$ & $\underline{0.32} {\scriptstyle \pm 0.03}$ & $0.0047 {\scriptstyle \pm 0.0006}$ & $0.22 {\scriptstyle \pm 0.03}$ & $1.74 {\scriptstyle \pm 0.10}$ & $1.24 {\scriptstyle \pm 0.16}$ & $1.11 {\scriptstyle \pm 0.17}$ \\ 
                NHP   & $0.89 {\scriptstyle \pm 0.09}$ & $0.53 {\scriptstyle \pm 0.24}$ & $\mathbf{0.0022} {\scriptstyle \pm 0.0007}$ & $0.31 {\scriptstyle \pm 0.12}$ & $2.00 {\scriptstyle \pm 0.30}$ & $1.30 {\scriptstyle \pm 0.26} $ & $ \underline{1.03} {\scriptstyle \pm 0.35}$  \\
            Diffusion & $\mathbf{0.61} {\scriptstyle \pm 0.10}$ & $0.33 {\scriptstyle \pm 0.04}$ & $\underline{0.0037} {\scriptstyle \pm 0.0012}$ & $0.23 {\scriptstyle \pm 0.14}$ & $\mathbf{1.30} {\scriptstyle \pm 0.21}$ & $\mathbf{0.86} {\scriptstyle \pm 0.18}$ & $\mathbf{0.92} {\scriptstyle \pm 0.20}$ \\
         Add-and-Thin & $0.86 {\scriptstyle \pm 0.05}$ & $\mathbf{0.30} {\scriptstyle \pm 0.04}$ & $0.0043 {\scriptstyle \pm 0.0007}$ & $\underline{0.21} {\scriptstyle \pm 0.03}$ & $\underline{1.53} {\scriptstyle \pm 0.14} $ & $1.16 {\scriptstyle \pm 0.16} $ & $1.20 {\scriptstyle \pm 0.14}$  \\
         EventFlow {\scriptsize (25 NFEs)} & $\underline{0.68} {\scriptstyle \pm 0.02}$& $1.06 {\scriptstyle \pm 0.09}$& $0.028 {\scriptstyle \pm 0.0015}$ & $\mathbf{0.19} {\scriptstyle \pm 0.01}$ & $2.30 {\scriptstyle \pm 0.22}$& $\underline{0.90} {\scriptstyle \pm 0.05 }$& $1.30 {\scriptstyle \pm 0.03}$ \\ \midrule
         EventFlow {\scriptsize (10 NFEs)} & ${0.60} {\scriptstyle \pm 0.02}$& $1.01 {\scriptstyle \pm 0.31}$& $0.0113 {\scriptstyle \pm 0.0016}$& ${0.16} {\scriptstyle \pm 0.01}$& $1.93 {\scriptstyle \pm 0.07}$& ${0.77} {\scriptstyle \pm 0.05}$& ${1.14} {\scriptstyle \pm 0.07}$ \\
         EventFlow {\scriptsize (1 NFE)} & $0.65 {\scriptstyle \pm 0.16}$& $1.30 {\scriptstyle \pm 0.61}$& $0.0445{\scriptstyle \pm 0.0104}$& $0.18 {\scriptstyle \pm 0.04}$& $1.52 {\scriptstyle \pm 0.17}$& $0.95 {\scriptstyle \pm 0.17}$& $1.03 {\scriptstyle \pm 0.24}$\\ \bottomrule
 \end{tabular}
}
\end{table}

\FloatBarrier
\subsection{Forecasting TPPs} 
In Table \ref{tab:forecast}, we report the sequence distance values appearing in the forecasting experiment (i.e., Figure \ref{fig:sequence_distance_results} and Table \ref{tab:forecast_ablation} in the main paper) with standard deviations. We additionally report an ablation where we use the true value of $n$, rather than sampling $n \sim p_\phi(n \mid \HH)$. This serves to measure how much room for improvement remains from the event count predictor. In general, we see that this benchmark is still not saturated, with further gains being possible especially on datasets with long sequences (e.g., the Reddit-C and Reddit-S datasets).

\begin{table}[!b]
\vspace{-1cm}
\caption{Sequence distance \eqref{eqn:sequence_distance} between the forecasted and ground-truth event sequences on a held-out test set. Lower is better. We report the mean $\pm$ one standard deviation over five random seeds. The lowest mean distance on each dataset is indicated in bold, and the second lowest by an underline.}
\label{tab:forecast}
\centering
\small{
\begin{tabular}{@{}lrrrrrrr@{}}
\toprule
                        & PUBG & Reddit-C & Reddit-S & Taxi & Twitter & Yelp-A & Yelp-M \\ \midrule
                IFTPP & $4.2 {\scriptstyle \pm 0.7}$ & $25.6 {\scriptstyle \pm 2.3}$ & $61.2 {\scriptstyle \pm 3.2}$ & $5.1 {\scriptstyle \pm 0.4}$ & $2.9 {\scriptstyle \pm 0.2}$ & $2.1 {\scriptstyle \pm 0.2}$ & $3.4 {\scriptstyle \pm 0.2}$ \\
                  NHP & $2.8 {\scriptstyle \pm 0.1}$ & $31.0 {\scriptstyle \pm 1.4}$ & $95.7 {\scriptstyle \pm 0.7}$ & $4.5 {\scriptstyle \pm 0.3}$ & $3.4 {\scriptstyle \pm 0.5}$ & $\underline{1.8} {\scriptstyle \pm 0.1}$ & $\underline{3.0} {\scriptstyle \pm 0.2}$    \\
            Diffusion & $5.4 {\scriptstyle \pm 1.2}$ & $25.7 {\scriptstyle \pm 0.9}$ & $80.3 {\scriptstyle \pm 11.4}$ & $4.6 {\scriptstyle \pm 0.7}$ & $\underline{2.4} {\scriptstyle \pm 0.2}$ & $\underline{1.8} {\scriptstyle \pm 0.1}$ & $3.3 {\scriptstyle \pm 0.7}$ \\
         Add-and-Thin & $\underline{2.5} {\scriptstyle \pm 0.04}$ & $\underline{22.2} {\scriptstyle \pm 4.6}$ & $\underline{34.3} {\scriptstyle \pm 0.4}$ & $\underline{3.7} {\scriptstyle \pm 0.1}$ & $3.1 {\scriptstyle \pm 0.2}$ & $\underline{1.8} {\scriptstyle \pm 0.1}$ & $\underline{3.0} {\scriptstyle \pm 0.2}$ \\
EventFlow {\scriptsize (25 NFEs)} & $\mathbf{2.0} {\scriptstyle \pm 0.03}$ & $\mathbf{15.8} {\scriptstyle \pm 2.7}$& $\mathbf{16.0} {\scriptstyle \pm 0.2}$& $\mathbf{3.2} {\scriptstyle \pm 0.1}$& $\mathbf{1.4} {\scriptstyle \pm 0.01}$ & $\mathbf{1.3} {\scriptstyle \pm 0.02}$ & $\mathbf{1.9} {\scriptstyle \pm 0.1}$ \\ \midrule
EventFlow {\scriptsize (10 NFEs)} & $2.0 {\scriptstyle \pm 0.03}$& $15.8 {\scriptstyle \pm 2.7}$& $15.8 {\scriptstyle \pm 0.2}$& $3.1 {\scriptstyle \pm 0.1}$ & $1.3 {\scriptstyle \pm 0.02}$ & $1.3 {\scriptstyle \pm 0.1}$ & $1.9 {\scriptstyle \pm 0.1}$ \\
EventFlow {\scriptsize (1 NFE)} & $2.0 {\scriptstyle \pm 0.01}$& $15.8 {\scriptstyle \pm 2.7}$& $15.8 {\scriptstyle \pm 0.2}$& $3.2 {\scriptstyle \pm 0.2}$& $1.4 {\scriptstyle \pm 0.03}$ & ${1.8 \scriptstyle \pm 0.3}$& $1.9 {\scriptstyle \pm 0.1}$ \\
EventFlow {\scriptsize (25 NFEs, true $n$)} & $1.2 {\scriptstyle \pm 0.01}$& $5.5 {\scriptstyle \pm 0.3}$& $8.8 {\scriptstyle \pm 0.2}$& $1.8 {\scriptstyle \pm 0.02}$& $0.7 {\scriptstyle \pm 0.01}$& $0.7 {\scriptstyle \pm 0.02}$&$1.1 {\scriptstyle \pm 0.02}$\\ \bottomrule
\end{tabular}
}
\end{table}

\subsection{Runtimes}
In Table~\ref{tab:runtimes}, we report the wall-clock time required to generate $1\,000$ sequences for each method. As expected, NHP and IFTPP achieve the fastest generation times, though this comes at the cost of substantially lower forecast quality (see Figure~\ref{fig:sequence_distance_results}). Both Diffusion and Add-and-Thin are relatively slower due to their iterative refinement procedures. EventFlow (with 1 NFE) attains state-of-the-art forecasting accuracy while maintaining generation times comparable to IFTPP on datasets with moderate sequence lengths (Yelp-A, Yelp-M, Taxi). On datasets with longer sequences (e.g., Reddit-S), the transformer-based architecture of EventFlow naturally incurs higher computational cost. Although EventFlow with 25 NFEs is slower, Table~\ref{tab:forecast_ablation} shows that a single NFE already suffices to reach SOTA performance. We note that these runtime comparisons are meant as approximate indicators, since implementation and architectural details can substantially influence wall-clock times. Overall, EventFlow achieves competitive generation efficiency despite not being explicitly optimized for speed.

\begin{table}[!h]
\caption{Wall-clock runtimes (seconds) required to generate $1\,000$ sequences from each model, at the largest batch size which fit in memory.}
\vspace{0.5em}
\label{tab:runtimes}
\centering
\small{
\begin{tabular}{@{}lrrrrrrr@{}}
\toprule
                        & PUBG & Reddit-C & Reddit-S & Taxi & Twitter & Yelp-A & Yelp-M \\ \midrule
IFTPP & $16.15$ & $20.77$ & $69.14$ & $22.62$ & $1.85$ & $6.55$ & $9.60$ \\ 
NHP & $0.99$ & $6.13$ & $11.94$ & $1.19$ & $0.83$ & $0.74$ & $0.89$ \\
Diffusion & $43.77$ & $99.82$ & $262.26$ & $78.77$ & $43.41$ & $29.82$ & $32.85$ \\
Add-and-Thin & $149.09$ & $261.53$ & $292.51$ & $42.60$ & $16.71$ & $34.73$ & $34.87$ \\
EventFlow (1 NFE) & $35$ & $739$ & $885$ & $8$ & $42$ & $6$ & $10$ \\
EventFlow (25 NFEs) & $769$ & $17889$ & $19624$ & $187$ & $960$ & $142$ & $233$ \\ \bottomrule

\end{tabular}
}
\end{table}

\end{document}